\documentclass{article} 

\usepackage{iclr2025_conference,times}

\usepackage{wrapfig}
\usepackage{lipsum} 
\usepackage{wrapfig}

\usepackage{booktabs}      
\usepackage{multirow}      
\usepackage{amssymb}       
\usepackage{graphicx, subfigure}     

\usepackage{colortbl}
\usepackage{xcolor}

\usepackage{hyperref}

\usepackage{url}
\usepackage{multicol}
\usepackage{bm}
\usepackage{aliascnt}
\usepackage{adjustbox}
\usepackage{siunitx} 
\usepackage{booktabs}
\usepackage{multirow} 
\usepackage{enumitem}
\usepackage{booktabs} 
\usepackage{amssymb}  
\usepackage{amsmath}  
\usepackage{graphicx} 
\usepackage{longtable} 
\usepackage{graphicx}
\usepackage{subcaption}
\usepackage{calc}
\usepackage[most]{tcolorbox}
\usepackage{listings}
\usepackage{wrapfig}
\usepackage{caption}
\usepackage{graphicx}

\definecolor{uclablue}{rgb}{0.15, 0.45, 0.68}
\hypersetup{
    breaklinks,
    citecolor=uclablue,
    colorlinks=true,
}

\newtcolorbox{AIbox}[2][]{aibox,title=#2,#1}
\tcbset{
  aibox/.style={
    top=10pt,
    colframe=black,
    colbacktitle=black,
    coltitle=white,
    center,
  }
}

\lstdefinelanguage{prompt}{
    basicstyle=\scriptsize\ttfamily, 
    mathescape=true,        
    escapebegin=\color{latentcolor},  
    escapeend={},
    escapechar=@,
    stringstyle = \color{myorange},
    showstringspaces = false,
    moredelim = [s][\color{mypink}]{`}{`},
    moredelim = [s][\color{mybrown}]{```json}{```},
    moredelim = [s][\color{latentcolor}]{<StartOfLatent>}{<EndOfLatent>},
    literate = %
        {\ \ a.\ }{{\textcolor{mypurple}{\ \ a.\ }}}5
        {\ \ b.\ }{{\textcolor{mypurple}{\ \ b.\ }}}5
        {\ \ c.\ }{{\textcolor{mypurple}{\ \ c.\ }}}5
        {\ \ d.\ }{{\textcolor{mypurple}{\ \ d.\ }}}5
        {\ \ e.\ }{{\textcolor{mypurple}{\ \ e.\ }}}5
        {\ \ f.\ }{{\textcolor{mypurple}{\ \ f.\ }}}5
        {\ \ g.\ }{{\textcolor{mypurple}{\ \ g.\ }}}5
        {\ \ h.\ }{{\textcolor{mypurple}{\ \ h.\ }}}5
        {\ I.\ }{{\textcolor{mypurple}{\ I.\ }}}4
        {\ II.\ }{{\textcolor{mypurple}{\ II.\ }}}5
        {\ III.\ }{{\textcolor{mypurple}{\ III.\ }}}6
        {\ IV.\ }{{\textcolor{mypurple}{\ IV.\ }}}5
        {\ V.\ }{{\textcolor{mypurple}{\ V.\ }}}4
}

\newtcblisting[auto counter, number within=subsection]{prompt}[4][]{%
  before skip=6pt, after skip=6pt,    
  top=6pt, bottom=6pt, left=7pt, right=7pt, 
  enhanced, breakable,
  colback=white, colframe=gray!65,
  boxrule=0.6pt, arc=2pt,
  drop shadow={black!15},             
  listing only,
  listing options={
    language=prompt,
    upquote=true,
    basicstyle=\scriptsize\ttfamily \setlength{\baselineskip}{1.1\baselineskip},
    columns=fullflexible,
    breaklines=true,
    breakindent=0pt,
    xleftmargin=\ifx\empty#2\empty0pt\else#2\fi,
    xrightmargin=\ifx\empty#3\empty0pt\else#3\fi,
    aboveskip=0pt,                    
    belowskip=0pt,                    
    showstringspaces=false,
  },
  title={\ifstrempty{#4}{Prompt \thetcbcounter}{Prompt \thetcbcounter: #4}},
  colbacktitle=gray!6, coltitle=black,
  attach boxed title to top left={yshift=-1mm, xshift=6pt},
  boxed title style={colback=gray!6, boxrule=0pt, sharp corners},
  #1
}

\newtcblisting[auto counter, number within=subsection]{showcase}[4][]{%
  before=\par\vspace{\baselineskip},
  after=\par,
  width=\linewidth,
  enhanced,
  arc=0em,
  boxrule=1pt,
  listing only,
  listing options={
    language=prompt,
    upquote=true,
    basicstyle=\scriptsize\ttfamily \setlength{\baselineskip}{1.1\baselineskip},
    breaklines=true,
    breakindent=0pt,
    xleftmargin=\ifx\empty#2\empty-12pt\else#2\fi,
    xrightmargin=\ifx\empty#3\empty-5pt\else#3\fi,
    aboveskip=-4pt,
    belowskip=-4pt,
    columns=fullflexible,
    },
  colback=white,
  colframe=gray,
  colbacktitle=gray!5,
  coltitle=black,
  attach boxed title to top center={yshift=-3mm},
  box align=center,
  parbox=false,
  title={\ifstrempty{#4}{Example \thetcbcounter}{Example \thetcbcounter: #4}},
  #1
}

\definecolor{linkColor}{rgb}{0.2,0.4,0.6}
\definecolor{myblue}{HTML}{0379AC}
\definecolor{myred}{HTML}{A50E50}
\definecolor{myorange}{RGB}{238, 133, 74}
\definecolor{latentcolor}{named}{cyan}
\definecolor{normalcolor}{RGB}{0, 0, 0}
\usepackage{marvosym}
\definecolor{lightblue1}{rgb}{0.97, 0.985, 1} 
\definecolor{lightblue2}{rgb}{0.92, 0.965, 1} 
\definecolor{lightblue3}{rgb}{0.84, 0.93, 1}
\definecolor{lightblue4}{rgb}{0.74, 0.87, 1}
\definecolor{lightblue5}{rgb}{0.64, 0.81, 1}
\definecolor{lightblue6}{rgb}{0.54, 0.75, 1}

\definecolor{lightgreen1}{rgb}{0.97, 1.00, 0.97}
\definecolor{lightgreen2}{rgb}{0.92, 0.98, 0.92}
\definecolor{lightgreen3}{rgb}{0.84, 0.95, 0.84}
\definecolor{lightgreen4}{rgb}{0.74, 0.91, 0.74}
\definecolor{lightgreen5}{rgb}{0.64, 0.86, 0.64}
\definecolor{lightgreen6}{rgb}{0.54, 0.81, 0.54}

\definecolor{lightorange1}{rgb}{1.00, 0.98, 0.95}
\definecolor{lightorange2}{rgb}{1.00, 0.95, 0.85}
\definecolor{lightorange3}{rgb}{1.00, 0.90, 0.70}
\definecolor{lightorange4}{rgb}{1.00, 0.85, 0.55}
\definecolor{lightorange5}{rgb}{1.00, 0.80, 0.40}
\definecolor{lightorange6}{rgb}{1.00, 0.75, 0.30}

\definecolor{lightpurple1}{rgb}{0.985, 0.97, 1.00}
\definecolor{lightpurple2}{rgb}{0.96, 0.92, 1.00}
\definecolor{lightpurple3}{rgb}{0.93, 0.84, 1.00}
\definecolor{lightpurple4}{rgb}{0.87, 0.74, 1.00}
\definecolor{lightpurple5}{rgb}{0.81, 0.64, 1.00}
\definecolor{lightpurple6}{rgb}{0.75, 0.54, 1.00}

\definecolor{lightred1}{rgb}{1.00, 0.97, 0.97}
\definecolor{lightred2}{rgb}{1.00, 0.92, 0.92}
\definecolor{lightred3}{rgb}{1.00, 0.84, 0.84}
\definecolor{lightred4}{rgb}{1.00, 0.74, 0.74}
\definecolor{lightred5}{rgb}{1.00, 0.64, 0.64}
\definecolor{lightred6}{rgb}{1.00, 0.54, 0.54}

\definecolor{lightcyan1}{rgb}{0.97, 1.00, 1.00}
\definecolor{lightcyan2}{rgb}{0.92, 0.98, 0.98}
\definecolor{lightcyan3}{rgb}{0.84, 0.95, 0.96}
\definecolor{lightcyan4}{rgb}{0.74, 0.91, 0.94}
\definecolor{lightcyan5}{rgb}{0.64, 0.87, 0.92}
\definecolor{lightcyan6}{rgb}{0.54, 0.83, 0.90}

\usepackage{xcolor,colortbl}
\definecolor{Gray}{gray}{0.85}
\definecolor{LightCyan}{rgb}{0.88,1,1}
\definecolor{greyC}{RGB}{180,180,180}
\definecolor{greyL}{RGB}{235,235,235}
\definecolor{citeColor}{RGB}{0,20,115}
\hypersetup{colorlinks,linkcolor={red},citecolor={citeColor},urlcolor={citeColor}}
\definecolor{shadecolor}{rgb}{0.92,0.92,0.92}

\usepackage{graphicx}
\usepackage{subcaption}
\usepackage{url}

\usepackage{booktabs}
\usepackage{multirow}
\usepackage{cleveref}
\crefname{template}{Template}{Template}

\usepackage{enumitem} 
\usepackage[most]{tcolorbox}
\usepackage{pifont}
\definecolor{rliableblue}{RGB}{0, 102, 204} 

\tcbset{
    myblueboxstyle/.style={
        colback=lightblue3, 
        colframe=lightblue5, 
        coltitle=black, 
        fonttitle=\bfseries, 
        boxrule=1pt, 
        width=\linewidth, 
        left=2pt, 
        right=2pt, 
        top=2pt, 
        bottom=2pt, 
        sharp corners, 
        boxsep=2pt
    }
}

\lstdefinestyle{iclrstyle}{
    language=Python,
    basicstyle=\ttfamily\small,  
    columns=fullflexible,        
    keepspaces=true,             
    showspaces=false,            
    showstringspaces=false,      
    commentstyle=\color{gray}\itshape, 
    keywordstyle=\color{codekw}\bfseries, 
    stringstyle=\color{myorange}, 
    escapechar=|,                
    frame=none,                  
    xleftmargin=1.5em,           
    aboveskip=0.5em,             
    belowskip=0.5em,             
    breaklines=true,             
    breakindent=0pt,
}

\usepackage{graphicx}
\usepackage{booktabs}
\usepackage{pifont}
\usepackage[table]{xcolor}  
\usepackage{wrapfig}
\usepackage{multirow}
\usepackage{float}
\usepackage{pifont}
\usepackage{amssymb}

\usepackage[most]{tcolorbox}
\usepackage{multicol}
\usepackage{tcolorbox}
\usepackage{titlesec}
\definecolor{objblue}{RGB}{3,139,221}  
\definecolor{attrred}{RGB}{255,67,67}    
\definecolor{easygreen}{RGB}{0,156,75}  
\definecolor{middleyellow}{RGB}{242,89,34}  
\definecolor{hardred}{RGB}{216,56,58}
\usepackage{colortbl}
\usepackage{array}
\definecolor{BoxBackground}{RGB}{240, 240, 240} 
\definecolor{BoxFrame}{RGB}{0, 0, 0} 
\definecolor{TitleBackground}{RGB}{0, 0, 0} 
\definecolor{TitleText}{RGB}{255, 255, 255} 
\tcbset{
  academicbox/.style={
    boxsep=5pt,
    left=2pt,
    right=2pt,
    bottom=0.5pt,
    boxrule=0.5pt,
    colback=BoxBackground,
    colframe=BoxFrame,
    colbacktitle=TitleBackground,
    coltitle=TitleText,
    enhanced,
    attach boxed title to top left={yshift=-0.1in,xshift=0.1in},
    boxed title style={boxrule=0pt,colframe=white},
    title={#1},
  }
}
\newtcolorbox{AcademicBox}[1][]{academicbox=#1}

\usepackage{algorithm}
\usepackage{algpseudocode}
\usepackage{listings} 
\usepackage{xcolor}   
\usepackage{etoolbox}
\makeatletter
\AfterEndEnvironment{algorithm}{\let\@algcomment\relax}
\AtEndEnvironment{algorithm}{\kern2pt\hrule\relax\vskip3pt\@algcomment}
\let\@algcomment\relax
\newcommand\algcomment[1]{\def\@algcomment{\footnotesize#1}}
\renewcommand\fs@ruled{\def\@fs@cfont{\bfseries}\let\@fs@capt\floatc@ruled
  \def\@fs@pre{\hrule height.8pt depth0pt \kern2pt}%
  \def\@fs@post{}%
  \def\@fs@mid{\kern2pt\hrule\kern2pt}%
  \let\@fs@iftopcapt\iftrue}
\makeatother

\NewDocumentCommand{\xx}
{ mO{} }{\textcolor{blue}{\textsuperscript{\textit{todo}}\textsf{\textbf{\small[#1]}}}}
\usepackage{xspace}

\usepackage{soul}
\definecolor{codeblue}{rgb}{0.25,0.5,0.5}
\definecolor{codekw}{rgb}{0.85, 0.18, 0.50}
\definecolor{diffgreen}{rgb}{0.0, 0.6, 0.0} 
\definecolor{diffred}{rgb}{0.8, 0.0, 0.0}   

\title{LoomVideo: Unifying Multimodal Inputs into \\ Video Generation and Editing} 

\author{
\vspace{-0.5em}
\centering
\textbf{\large Peking University, Alibaba Group} \\[1.2em]
{\small
\raisebox{0pt}{\includegraphics[height=1.0em]{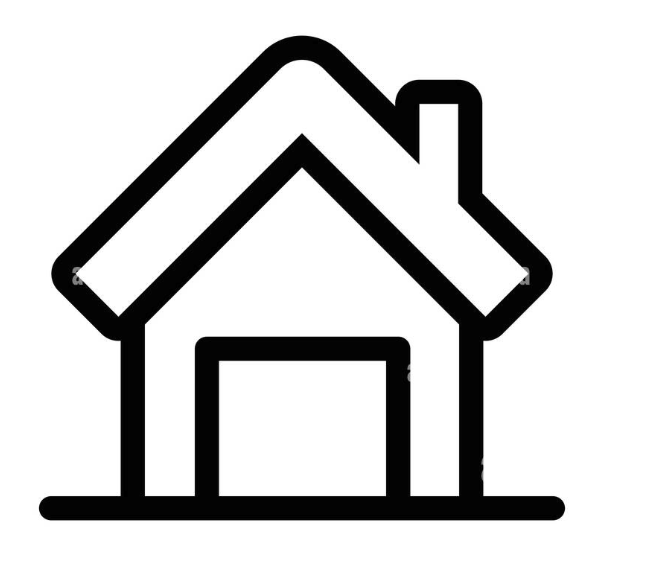}\hspace{0.3em}} 
Project Page: \url{https://msalab-pku.github.io/projects/LoomVideo/index.html} \\
\raisebox{0pt}{\includegraphics[height=1.0em]{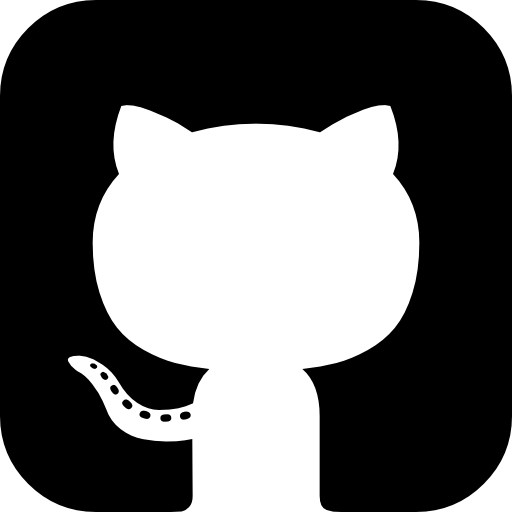}\hspace{0.3em}} 
Github: \url{https://github.com/MSALab-PKU/LoomVideo} \\
\raisebox{0pt}{\includegraphics[height=1.0em]{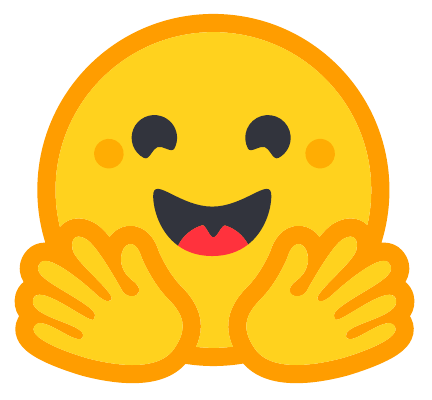}\hspace{0.3em}} 
Model: \url{https://huggingface.co/MSALab/LoomVideo}
}
}

\begin{document}
\maketitle
\renewcommand*{\thefootnote}{\fnsymbol{footnote}}

\begin{abstract}
Developing unified video generation and editing models capable of interpreting interleaved multimodal inputs is a promising yet challenging frontier field. Existing unified frameworks predominantly rely on massive models (typically 13B parameters or more) and incorporate source video conditions for editing by concatenating sequence tokens. This concatenation inevitably doubles the sequence length, quadrupling the computational complexity of the self-attention mechanism and introducing prohibitive overhead. To address these bottlenecks, we present \textbf{LoomVideo}, a highly efficient 5B-parameter unified architecture for both video generation and editing. LoomVideo replaces the standard text encoder with a Multimodal Large Language Model (MLLM) and employs a \textbf{Deepstack} injection mechanism to align multi-layer MLLM features with the Diffusion Transformer (DiT). Crucially, we introduce a zero-overhead \textbf{Scale-and-Add} conditioning approach for video editing. By scaling and directly adding the clean source video latent to the noised target latent, this elegant design eliminates the need for token concatenation, drastically reducing computational cost while maintaining robust capabilities for complex, non-rigid edits. Furthermore, a \textbf{Negative Temporal RoPE} strategy is seamlessly integrated to handle multiple reference images. Extensive experiments demonstrate that our compact 5B model achieves state-of-the-art or highly competitive performance across comprehensive benchmarks, exhibiting exceptional superiority in e-commerce and fashion generation scenarios. Benefiting from the zero-overhead conditioning mechanism, LoomVideo achieves at least a 5.41$\times$ acceleration in inference speed compared to models of similar capabilities, paving the way for highly practical and efficient video foundation models.
\end{abstract}
\section{Introduction}

The demand for versatile and highly controllable video generation and editing is rapidly increasing across diverse practical applications, from digital entertainment to e-commerce. To meet these complex requirements, developing a unified architecture capable of interpreting and adhering to interleaved multimodal inputs (e.g., video, images, and fine-grained instructions) is of paramount importance. The paradigm of unifying multiple vision-language tasks within a single foundational model first achieves remarkable success in the image domain, evidenced by pioneering works such as Qwen-Image \cite{Qwen-Image} and the OmniGen series \cite{OmniGen, OmniGen2}. Recently, this unified modeling approach has been extended to the video domain, though existing works remain relatively sparse due to the inherent complexity of spatiotemporal dynamics.

A few pioneering works have explored unified video generation and editing, including UniVideo \cite{UniVideo}, OmniWeaving \cite{OmniWeaving}, VINO \cite{VINO}, and OmniVideo \cite{OmniVideo, OmniVideo2}. While these frameworks have demonstrated the feasibility of multimodal conditioning, they predominantly suffer from two major limitations. First, most of these frameworks rely heavily on massive base models (typically 13B parameters or more), making them exceptionally resource-intensive. Second, to incorporate source video condisource-video conditions for editing tasks, the prevailing practice is to concatenate the source-video tokens with the target-videosequence dimension. This design inevitably doubles the token sequence length, which in turn quadruples the computational complexity of the self-attention mechanism. Consequently, these approaches introduce prohibitive computational overhead and significantly elevate both training and inference costs.

To address these bottlenecks, we introduce \textbf{LoomVideo}, an efficient unified architecture for video generation and editing. Built upon the foundation of the 5B-parameter Wan 2.2 Text-Image-to-Video (TI2V) model~\cite{Wan}, we replace its standard T5 text encoder with Qwen3-VL~\cite{Qwen3-VL}, a powerful Multimodal Large Language Model (MLLM), to handle interleaved multimodal inputs. To maximize multimodal interaction and minimize computational overhead, we incorporate three key effective architectural designs. First, instead of solely utilizing the final-layer embedding of the vision-language model, we employ a \textbf{Deepstack} injection mechanism. 

\begin{figure}[H]
    \centering
    \includegraphics[width=\textwidth]{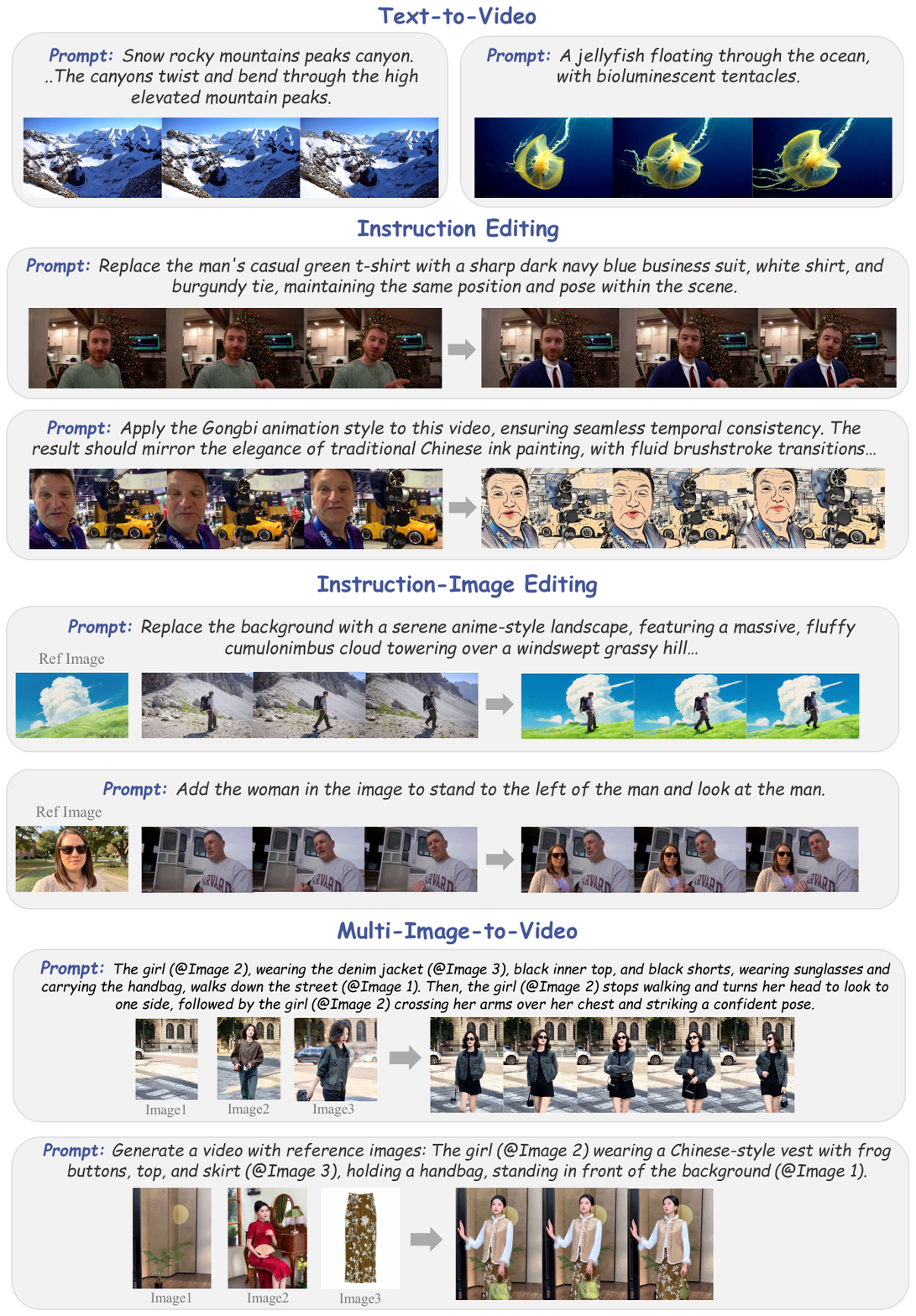} 
    \caption{Showcase of \textbf{LoomVideo} across diverse video generation and editing scenarios, including foundational video generation \& editing, reference-image-guided video generation and editing.}
    \label{fig:teaser}
\end{figure}

We extract feature embeddings from every layer of Qwen3-VL and inject them into the corresponding layers of the Diffusion Transformer (DiT) via cross-attention, thereby enhancing the deep semantic alignment between the multimodal controls and the generative process. Second, to overcome the severe efficiency bottleneck caused by token concatenation in editing tasks, we utilize a \textbf{Scale-and-Add} conditioning approach. Rather than appending source tokens, we simply scale the clean source video latent by the current timestep and add it directly to the noised target latent. This elegant design introduces zero additional tokens to the sequence length, drastically reducing generation time. More importantly, we demonstrate that this lightweight conditioning mechanism is fully capable of driving highly complex, non-rigid editing operations (e.g., changing human actions or camera angles). Third, to seamlessly integrate multiple reference images, we employ a \textbf{Negative Temporal RoPE index} strategy. This positional encoding effectively distinguishes the reference inputs from the video frames, providing robust multi-image guidance without disrupting the spatiotemporal dynamics of the generative process.

We devise a progressive three-stage training strategy: \textit{Stage 1} focuses on semantic alignment at a low resolution, where we extract and inject MLLM embeddings into the DiT. \textit{Stage 2} scales up the resolution, training simultaneously on fundamental image/video generation and reconstruction/editing tasks. \textit{Stage 3} incorporates diverse reference images and advanced text instructions to drive complex generation and editing tasks. Additionally, we further enhance the model's overall performance via reinforcement learning post-training, which effectively improves its instruction-following capability and generation fidelity. Through this progressive multimodal alignment, task adaptation, and targeted post-training, our relatively compact 5B model efficiently performs various generation and editing tasks that previously required 13B-scale models, unifying them within a single framework.

In summary, the main contributions of this work are as follows:
\begin{itemize}
    \item We propose \textbf{LoomVideo}, a highly efficient 5B-parameter unified architecture that seamlessly integrates visual-language multimodal input controls for both video generation and editing.
    \item Our method achieves state-of-the-art (SOTA) or on-par SOTA performance across multiple comprehensive benchmarks. It demonstrates particularly superior capabilities in reference-image-guided video editing and controllable generation tasks within e-commerce and fashion product domains.
    \item Benefiting from the compact parameter scale and the zero-overhead \textbf{Scale-and-Add} conditioning mechanism, LoomVideo achieves at least 5.41$\times$ acceleration in inference speed compared to existing concatenation-based unified models of similar capabilities.
\end{itemize}
\section{Related Work}
\label{sec:related-work}

\textbf{Video Generation and Editing.}
The landscape of text-to-video generation has been reshaped by diffusion models. This evolution has catalyzed a wave of highly popular and powerful foundational video models, such as CogVideoX~\cite{CogVideoX}, Wan~\cite{Wan}, and the HunyuanVideo series~\cite{HunyuanVideo, HunyuanVideo1.5}, which have garnered widespread attention and extensive open-source community support. Recently, state-of-the-art models like Seedance 2.0~\cite{Seedance2.0}, SkyReelsV4~\cite{SkyReels-V4}, and Sora 2~\cite{Sora2} have pushed the boundaries of visual quality, physical world complexity, and generation duration.
Building upon robust video generation priors, video editing has naturally emerged as the next frontier. Early video editing methods predominantly focused on zero-shot adaptations or rigid structural preservation, which often struggled with complex, non-rigid transformations. The introduction of instruction-following and unified frameworks has significantly advanced this field. Models like InsViE~\cite{InsViE}, Ditto~\cite{Ditto}, OpenVE~\cite{OpenVE}, and Kiwi-Edit~\cite{Kiwi-Edit} have demonstrated the feasibility of performing diverse editing tasks via instruction and reference guidance. Despite these remarkable strides, most existing unified models either rely on massive parameter scales (13B or more). In contrast to these approaches, our work demonstrates that a relatively compact 5B-scale Diffusion Transformer (coupled with an 8B VLM) can achieve highly efficient and high-quality multi-task unification. Particularly in complex e-commerce and fashion generation scenarios.

\textbf{Multimodal Input for Video Generation.}
As the demand for precise control over video generation and editing grows, the field is rapidly moving beyond pure text-to-video generation toward multimodal conditioning. Current prominent frameworks, such as VINO~\cite{VINO}, VACE~\cite{VACE}, OmniWeaving~\cite{OmniWeaving} , UniVideo~\cite{UniVideo}, and Omni-Video~\cite{OmniVideo, OmniVideo2}, have explored various ways to fuse interleaved multimodal inputs into video generation. For the input video, existing methods often concatenate the source video tokens with the target tokens~\cite{UniVideo, OmniWeaving}. While effective, this design inevitably leads to a drastic increase in token sequence length, resulting in severe computational bottlenecks.
Our architecture takes a more efficient approach to multimodal conditioning. 1) To achieve deep semantic alignment between the multimodal input and the generative prior, we do not merely use the final output of the VLM. Instead, we extract the feature embeddings from \textit{every layer} form the VLM and inject them into the corresponding layers of the DiT via cross-attention. 2) When incorporating source video conditions for editing tasks, we simply scale and add the clean source video latent to the target noised video latent. This elegant design introduces \textit{zero} additional tokens to the sequence length, ensuring optimal computational efficiency. Surprisingly, despite this lightweight conditioning scheme, our model is fully capable of executing complex, \textit{non-rigid} editing tasks (e.g., changing human actions or camera directions).

\section{Method}

\begin{figure}[t]
    \centering
    \includegraphics[width=\textwidth]{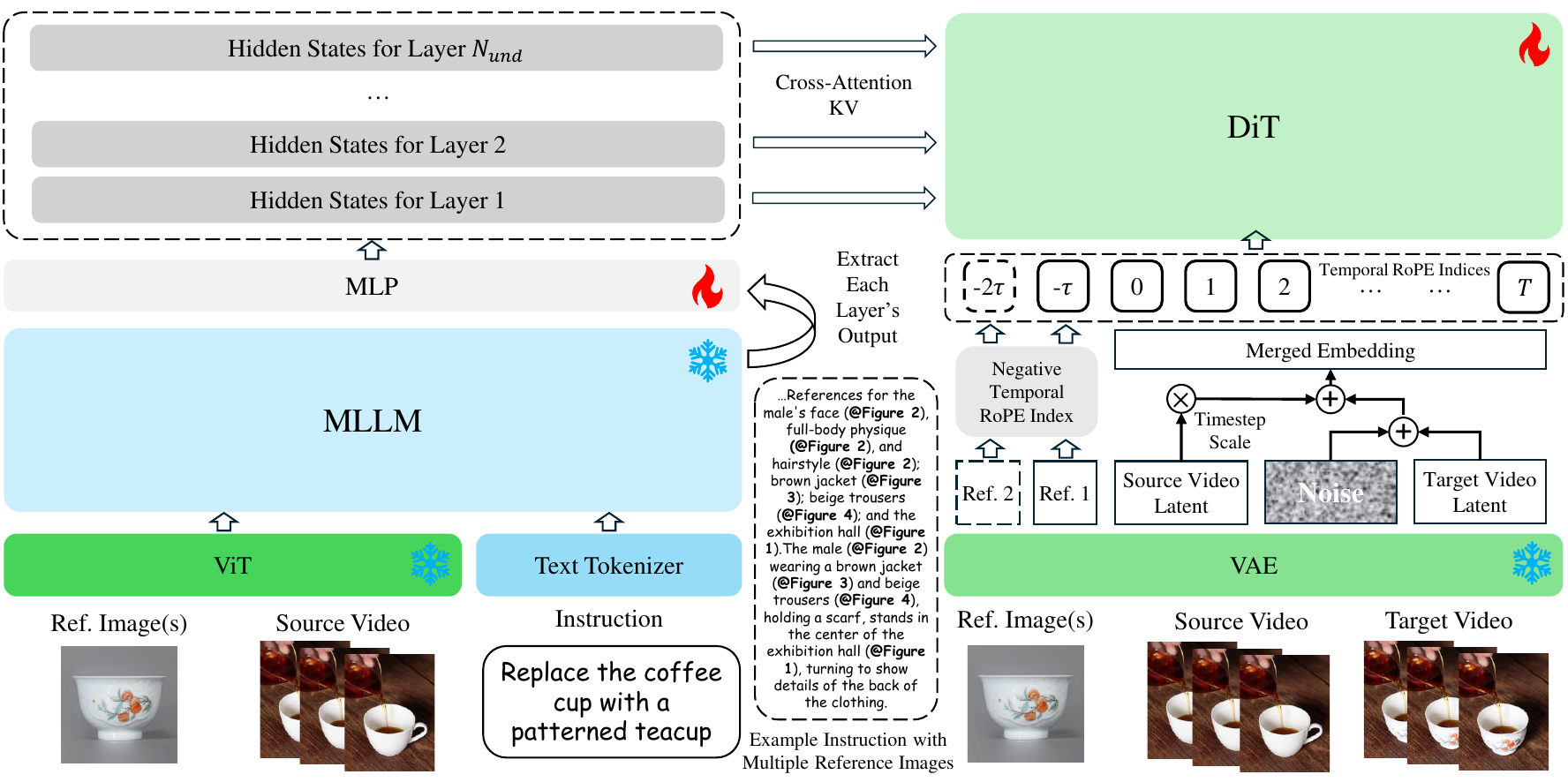} 
    \caption{\textbf{Overall Architecture of LoomVideo.} LoomVideo seamlessly processes interleaved multimodal inputs using an MLLM. It employs two key designs: (1) A \textbf{Deepstack} injection mechanism, which extracts feature embeddings from every layer of the MLLM and injects them into the corresponding layers of the DiT via cross-attention. (2) A zero-overhead \textbf{Scale-and-Add} conditioning approach for video editing, which scales the clean source video latent by the current timestep and directly adds it to the noised target latent, completely bypassing the severe inefficiency of token concatenation. (3) \textbf{Negative Temporal RoPE index} for multiple reference images.}
    \label{fig:architecture}
\end{figure}

\subsection{Architecture}

In this section, we introduce the architecture and key designs of LoomVideo, as illustrated in~\cref{fig:architecture}.

\textbf{Deepstack mechanism of MLLM features.}
\label{sec:deepstack}
To handle interleaved multimodal inputs (e.g., text, reference images, and source videos), we replace the default T5 text encoder with an MLLM. However, relying solely on the final-layer embedding of the MLLM restricts the generative prior from fully utilizing the rich, hierarchical semantic information captured at different layers. To maximize multimodal interaction, we propose a \textbf{Deepstack} injection mechanism. Rather than using only the final layer's output, we extract hidden states from every layer of the MLLM. These representations are transformed by a newly added MLP module and then injected into the corresponding layers of the DiT via cross-attention, serving as key and value. This process can be written as:
\begin{equation}
    \textbf{c}^{l} = \text{MLP}(\textbf{h}^{l}_{\text{mllm}}), \quad
    \textbf{o}^{l}_{\text{dit}} = \text{CrossAttn}(\textbf{h}^{l}_{\text{dit}}, \textbf{c}^{l}, \textbf{c}^{l}),
\end{equation}
where $\textbf{h}^{l}_{\text{mllm}})$ denotes the hidden states extracted from the $l$-th layer of the MLLM, and $\textbf{h}^{l}_{\text{dit}}$ represents the intermediate features of the corresponding $l$-th layer in the DiT. The MLP module projects the MLLM hidden states into the condition features $\textbf{c}^{l}$, which are then served as keys and values in the cross-attention module. The MLP module shares parameters across layers.
This layer-to-layer conditioning approach guarantees deep semantic alignment between the fine-grained multimodal controls and the spatiotemporal generative process, enhancing the model's instruction-following capabilities without introducing heavy adapter networks.

\textbf{Scale-and-Add of Source Video Latent.}
\label{sec:scale-and-add}
Existing unified video generation and editing frameworks predominantly incorporate source video conditions by concatenating the source video tokens with the target video tokens along the sequence dimension. This paradigm inevitably doubles the token sequence length, quadrupling the computational complexity of the self-attention mechanism and causing severe efficiency bottlenecks during both training and inference. To overcome this limitation, LoomVideo introduces a zero-overhead \textbf{Scale-and-Add} conditioning approach. Instead of appending source tokens, we directly manipulate the latent representations. Specifically, we scale the clean source video latent by a timestep-dependent factor and add it directly to the noised target latent at the current timestep. This merged embedding is then fed into the DiT. This process can be formulated as:
\begin{equation}
    \textbf{h}_{\text{merged}} = \phi(\textbf{z}_{\text{target}}) + t \cdot \phi'(\textbf{z}_{\text{source}}),
\end{equation}
where $\textbf{z}_{\text{target}}$ denotes the noised target video latent at the current timestep $t \in [0, 1)$, and $\textbf{z}_{\text{source}}$ represents the clean latent of the source video. Here, the timestep $t$ directly acts as the scaling factor. As $t$ increases, the scale factor proportionally amplifies the contribution of $\textbf{z}_{\text{source}}$, ensuring stronger guidance during the early stages of the denoising process, while leaving purer target latent during later steps. $\phi$ and $\phi'$ are the original and newly added patch embedding layers, respectively. $\phi'$ is zero-initialized. This lightweight conditioning introduces zero additional tokens to the sequence length, drastically reducing the generation time (achieving \textbf{at least 5.41$\times$} acceleration in inference speed compared to concatenation-based models). Despite its simplicity, empirical results in~\cref{sec:experiments} demonstrate that this mechanism provides sufficient guidance to drive highly complex, non-rigid editing operations, such as modifying human actions or camera angles, while preserving optimal computational efficiency.

\textbf{Negative Temporal RoPE Index for Reference Images.}
\label{sec:negative-rope}
Target video frames in DiT are assigned normal positive temporal RoPE index $0, 1, 2, \dots, T$, where $T$ is the number of latent frames. For the reference images, we assign them negative temporal indices $-\tau, -2\tau, \dots, -N_{ref}\tau$, where $\tau$ is a hyperparameter, and $N_{ref}$ is the number of reference images. By doing so, the model robustly distinguishes the reference images from the target video frames, providing strong multi-image guidance with minimal extra tokens. Furthermore, when the task involves detailed textual descriptions of multiple reference images (e.g., ``the person in @Figure 1", ``the clothes in @Figure 2"), this negative indexing strategy enables the model to explicitly align the $i$-th image specified in the prompt with its corresponding visual representation based on the absolute value of the assigned negative RoPE index.

\subsection{Training Data}
\label{sec:training_data}

We train LoomVideo using a mixture of image and video datasets. Although our primary objective is unified video generation and editing, incorporating a massive and semantically diverse collection of image generation and editing data effectively bolsters the foundational capabilities of the model.

\textbf{Text-to-Image/Video.} 
For text-to-image generation, we utilize approximately $\mathcal{O}(10\text{M})$ high-quality internal image-text pairs. Since we replace the default T5 text encoder with the Qwen3-VL MLLM, establishing accurate multimodal alignment from scratch is crutial. Our internal dataset features exceptionally detailed and accurate textual descriptions across a diverse range of visual domains, which significantly aids in maintaining robust text-image alignment performance. For text-to-video generation, we employ around $\mathcal{O}(10\text{M})$ open-source video-text pairs, primarily sourced from Koala 36M~\cite{Koala36M} and OpenVid-1M~\cite{OpenVid-1M}. This large-scale temporal data is sufficient to seamlessly extend the fundamental text-image alignment capabilities learned in the static domain to the video domain with complex spatiotemporal dynamics.

\textbf{Instruction-based Image/Video Editing.} 
For instruction-based image editing, we curate a comprehensive blend of open-source and internal datasets. The open-source collection comprises approximately $\mathcal{O}(10\text{M})$ samples from various high-quality datasets, including SEED-Data-Edit part 2 and 3~\cite{seed-data-edit}, NHR-Edit~\cite{NHR-Edit}, OmniGen2~\cite{OmniGen2}, Pico-Banana-400k~\cite{pico-banana-400k}, CrispEdit-2M~\cite{CrispEdit-2M}, ShareGPT-4o-Image~\cite{ShareGPT-4o-Image}, and GPT-Image-Edit-1.5M~\cite{GPT-Image-Edit-1.5M}. We further incorporate $\mathcal{O}(10\text{M})$ internal editing samples specifically focusing on the Taobao fashion domain. For instruction-based video editing, our training set consists of $\mathcal{O}(3\text{M})$ samples from the open-source Kiwi-Edit dataset~\cite{Kiwi-Edit}, supplemented by $\mathcal{O}(1\text{M})$ internal instruction-guided video editing pairs.

\textbf{Instruction-and-Reference-based Image/Video Editing.} 
To equip the model with fine-grained visual condition following capabilities, we incorporate reference-guided editing tasks. For the image domain, we predominantly rely on an internal dataset of $\mathcal{O}(10\text{M})$ instruction-and-reference pairs. For the video domain, we utilize approximately $\mathcal{O}(0.5\text{M})$ reference-guided video editing samples from the RefVIE dataset~\cite{Kiwi-Edit}, alongside $\mathcal{O}(1\text{M})$ internal samples tailored for reference-based video editing.

\textbf{Multiple-Reference-Image-to-Image/Video.} 
For tasks requiring multiple reference images, we utilize the open-source Phantom dataset~\cite{Phantom-Data} alongside a large-scale internal collection of multi-reference image and video data, particularly emphasizing the Taobao e-commerce domain. To ensure the model explicitly distinguishes and follows instructions involving multiple reference images, we adopt an interleaved text-image prompt format inspired by Seedance 2.0~\cite{Seedance2.0}. Specifically, we use explicit textual pointers such as ``The person in @Figure 1'' and ``The clothes in @Figure 2'' to bind textual descriptions with their corresponding visual reference inputs. An illustrative example of this interleaved multi-reference instruction format is provided in \cref{fig:architecture}.

\subsection{Training Pipeline}

\begin{table}[t]
\centering
\caption{Training recipe of LoomVideo.}
\label{tab:training-recipe}
\resizebox{0.8\linewidth}{!}{
\begin{tabular}{l|ccc}
\toprule
\multirow{2}{*}{\textbf{Hyperparameters}} & \textbf{Stage1} & \textbf{Stage2} & \textbf{Stage3} \\
& MLLM Alignment & Reconstruction \& Editing & Multi Task \\
\midrule
Learning rate & $2.0 \times 10^{-5}$ & $2.0 \times 10^{-5}$ & $2.0 \times 10^{-5}$ \\
LR scheduler & Constant & Constant & Constant \\
Weight decay & 0.0 & 0.0 & 0.0 \\
Gradient norm clip & 1.0 & 1.0 & 1.0 \\
Optimizer & AdamW & AdamW & AdamW \\
GPUs &  $16 \times$ H20s &  $64 \times$ H20s & $64 \times$ H20s \\
Batch Size & $\sim$ 640 & $\sim$ 128 & $\sim$ 128 \\
Warm-up steps & 100 & 1000 & 1000 \\
Training steps & 50K & 40K & 36K \\
EMA weight decay & None & 0.9999 & 0.9999 \\
EMA interval & None & 1 & 1 \\
Resolution & 256p & 480p & 480p \\
Text dropout ratio & 0 & 0.1 & 0.1 \\
All dropout ratio & 0 & 0.1 & 0.1 \\
Dynamic timestep shift scale & 0.65 & 0.5 & 0.5 \\
\midrule
\textbf{Data sampling ratio} & & & \\
Text-to-Image & 0.80 & 0.59 & 0.14 \\
Text-to-Video & 0.20 & 0.04 & 0.08 \\
Image Reconstruction & \cellcolor[HTML]{DCDCDC}0.00 & 0.07 & 0.02 \\
Video Reconstruction & \cellcolor[HTML]{DCDCDC}0.00 & 0.01 & 0.01 \\
Text-Image-to-Image & \cellcolor[HTML]{DCDCDC}0.00 & 0.28 & 0.17 \\
Text-Video-to-Video & \cellcolor[HTML]{DCDCDC}0.00 & 0.01 & 0.18 \\
Text-Image-Ref-to-Image & \cellcolor[HTML]{DCDCDC}0.00 & \cellcolor[HTML]{DCDCDC}0.00 & 0.15 \\
Text-Video-Ref-to-Video & \cellcolor[HTML]{DCDCDC}0.00 & \cellcolor[HTML]{DCDCDC}0.00 & 0.13 \\
Text-Multi-Image-to-Image & \cellcolor[HTML]{DCDCDC}0.00 & \cellcolor[HTML]{DCDCDC}0.00 & 0.06 \\
Text-Multi-Image-to-Video & \cellcolor[HTML]{DCDCDC}0.00 & \cellcolor[HTML]{DCDCDC}0.00 & 0.07 \\
\bottomrule
\end{tabular}
}
\end{table}

\textbf{Stage1: MLLM Alignment.} 
To enable our model to seamlessly process interleaved text, image, and video inputs, we must leverage the powerful multimodal understanding capabilities of the MLLM. The primary objective of the first stage is to replace the original condition encoder (i.e., the T5 text encoder) of the pretrained DiT with the MLLM. In this stage, we exclusively utilize text-to-image and text-to-video datasets. To ensure the generalizability of semantic alignment, we employ a large batch size (approximately $640$) while maintaining a relatively low spatial resolution of $256$p. The sample-level ratio of image to video data is set to 4:1. It is worth noting that since a single video contains significantly more tokens than an image, and we adopt a dynamic 1D sequence concatenation batching strategy bounded by a maximum token limit, the actual proportion of video tokens in a batch is substantially higher than that of image tokens. 

Furthermore, we ablated to retain the T5 text encoder, feeding both the T5 embeddings and the newly introduced MLLM embeddings into the DiT via cross-attention. Although this approach converged faster initially, we empirically observed in subsequent stages that the model heavily relied on the T5 embeddings and largely ignored the MLLM conditionings. Consequently, to force the generative prior to fully align with the MLLM representations, the T5 encoder must be completely discarded.

\textbf{Stage2: Reconstruction and Editing.} 
Following the first stage, the model demonstrates robust semantic alignment with the MLLM. However, its visual generation quality remains limited due to the low training resolution, often resulting in blurry details in intricate regions such as human faces. Therefore, the primary objective of the second stage is to scale up the resolution (to $480$p) and enhance generation quality. Simultaneously, we introduce a portion of instruction-based editing data to facilitate a smooth transition toward the multi-task training in the final stage. Importantly, we also introduce a reconstruction task utilizing the text-to-image and text-to-video datasets, prompted simply by ``reconstruct this image/video''. This design is motivated by the intuition that reconstruction serves as the fundamental prerequisite for editing; robust and accurate video editing capabilities must be built upon the model's ability to precisely reconstruct the source visual content.

\textbf{Stage3: Multi Task.} 
In the final stage, we incorporate the full spectrum of our datasets to perform comprehensive multi-task fine-tuning. During this phase, the data sampling distribution is deliberately skewed toward newly introduced and more challenging tasks, such as reference-guided image/video editing and multi-reference generation. This targeted sampling strategy ensures that the model devotes sufficient capacity to mastering complex, fine-grained control and non-rigid transformations.

\textbf{Post-training: Reinforcement Learning (RL).}
Current RL methods~\cite{flow-grpo, tp-grpo, PromptEcho,diffusionNFT} have demonstrated promising performance in visual generation tasks. We adopt DiffusionNFT~\cite{diffusionNFT} for our experiments, employing PickScore~\cite{pickscore} as the reward model to optimize human aesthetic preferences and perceptual quality. The training data is curated from our in-house dataset. We follow the FlowFactory~\cite{flow-factory} implementation for the training pipeline setup.

\subsection{Implementation Details}

\textbf{Model Initialization and Architecture.} 
We initialize our model with the pretrained weights of Wan 2.2 TI2V 5B~\cite{Wan} and Qwen3-VL-8B-Instruct~\cite{Qwen3-VL}. Because the Qwen3-VL-8B model consists of 36 transformer layers while the DiT contains only 30 layers, we extract the hidden states from the last 30 layers of the MLLM and inject them into the corresponding 30 layers of the DiT. To transform these MLLM hidden states for cross-attention, our newly introduced MLP module follows the architectural design of the original T5 text projection layer in DiT, which comprises two linear layers and an activation function. Notably, to ensure training stability, we prepend an RMSNorm layer at the beginning of this module to properly normalize the extracted Qwen3-VL hidden states before projection.

\textbf{Dynamic Batching and Multi-Resolution Training.} 
During training, we employ a 1D sequence concatenation batching strategy. Specifically, all spatial and temporal tokens from the images and videos within a batch are flattened into 1D sequences and concatenated together. We apply block-diagonal attention masks to strictly prevent tokens from different samples within the same batch from attending to one another. Under this paradigm, the batch size is dynamically constrained by a predefined maximum total token limit rather than a fixed number of samples, meaning the batch sizes reported in our settings are empirical averages. 
Furthermore, we adopt a dynamic multi-resolution training approach. Images and videos are grouped into predefined buckets based on their total pixel counts and aspect ratios, and are then resized to the nearest bucket resolution. By flattening and concatenating these dynamically sized samples, our framework natively supports efficient multi-resolution training.

\textbf{Dynamic Timestep Shift Scale.} 
Since our training paradigm involves a mixture of multi-resolution images and videos, we adopt the dynamic timestep shift technique recommended by Stable Diffusion 3 (SD3)~\cite{sd3}. This technique dynamically adjusts the noise scheduling by increasing the timestep shift alongside the growth of data resolution and temporal length. Specifically, the shift value is calculated as follows:
\begin{equation}
    \text{shift} = \text{scale} \times \log_2\left(\frac{N_{\text{target}}}{N_{\text{base}}}\right)
\end{equation}
where $N_{\text{target}}$ denotes the total token count of the target sample, and $N_{\text{base}}$ is the base token number. In our implementation, we set $N_{\text{base}} = 64$, which exactly corresponds to the sequence length of a standard $256 \times 256$ image after VAE encoding and patchification. Under this formulation, by setting the hyperparameter $\text{scale} = 0.5$, the calculated shift value for a high-resolution video (e.g., $480 \times 832 \times 97$) automatically reaches approximately $4.5$. This adaptive scaling ensures optimal noise scheduling across diverse input dimensions and modalities during joint training.

\textbf{Training Configurations.} 
Regarding the training resolutions listed in our recipe, ``256p'' denotes the bucket with a standard video dimension of $256 \times 256 \times 33$ frames (and its corresponding 1-frame image counterpart), while ``480p'' refers to a standard dimension of $480 \times 832 \times 97$ frames for videos (and corresponding 1-frame images). Finally, to maintain robust training stability and achieve better convergence, we maintain an Exponential Moving Average (EMA) of the model weights exclusively during Stage 2 and Stage 3. The detailed hyperparameters and specific data sampling ratios across all three stages are summarized in~\cref{tab:training-recipe}.

\section{Experiments}
\label{sec:experiments}

\subsection{Experimental Settings}

\textbf{Benchmarks.}
For the Text-to-Video (T2V) generation task, we employ VBench~\cite{VBench}, selecting Imaging Quality, Overall Consistency, and Subject Consistency as our primary evaluation metrics, and we report their average scores. For instruction-based video editing, we adopt OpenVE-Bench~\cite{OpenVE}. For instruction-and-reference-based video editing, we utilize RefVIE-Bench~\cite{Kiwi-Edit}. To assess multi-reference image-to-video (MI2V) capabilities, we evaluate our model on IntelligentVBench~\cite{OmniWeaving}, with a specific focus on its TIV2V and MI2V sub-tasks.

Furthermore, to comprehensively demonstrate our model's specialized capabilities in the e-commerce domain, we construct a novel benchmark, \textbf{FashionVideoBench}. Curated from internal, held-out data (strictly excluded from the training set), this benchmark encompasses six core tasks: Product Edit, Model Edit, Freeform Edit, Product-Reference Edit (PRef Edit), Model-Reference Edit (MRef Edit), and MI2V. Specifically, we sample 50 test cases for each sub-task, resulting in a total of 300 evaluation samples. We employ Gemini 2.5 Pro as an automated judge to assess the generated outputs across three dimensions: Subject Consistency, Prompt Following, and Video Quality. Detailed evaluation protocols and prompt templates are provided in Appendix \ref{app:3}.

\textbf{Baseline Methods.}
We primarily benchmark our approach against state-of-the-art open-source unified video generation and editing models, including UniVideo~\cite{UniVideo}, OmniWeaving~\cite{OmniWeaving}, VINO~\cite{VINO}, and VACE~\cite{VACE}. Depending on the task, we also compare against several specialized models. This includes editing-specific models such as Kiwi-Edit~\cite{Kiwi-Edit}, Ditto~\cite{Ditto}, OpenVE-Edit~\cite{OpenVE}, InsViE~\cite{InsViE}, and OmniVideo~\cite{OmniVideo}, as well as prominent foundational video generation models like Wan~\cite{Wan}, HunyuanVideo~\cite{HunyuanVideo, HunyuanVideo1.5}, CogVideoX~\cite{CogVideoX}, SkyReels-A2~\cite{SkyReels-A2}, SkyReels-V3~\cite{SkyReels-V3}, and Phantom~\cite{Phantom}. In addition, for certain benchmarks, we include comparisons with leading closed-source commercial models~\cite{Runway-Aleph, Kling-O1}. Importantly, most state-of-the-art models in this domain operate with massive parameter counts (typically 13B or larger), which are significantly larger than our efficient 5B-parameter architecture.

\subsection{Quantitative Comparison}

\begin{table*}[t]
    \centering
    \caption{Quantitative results on VBench.}
    \label{tab:quantitative-comparison-vbench}
    \resizebox{0.9\linewidth}{!}{
        \begin{tabular}{lc|ccc|c}
            \toprule
\textbf{Model} & \#Params & Imaging Quality & Overall Consistency & Subject Consistency & Average \\
\midrule

Wan 2.2 & 5B & \underline{69.82} & 22.41 & \underline{95.28} & {62.50} \\
UniVideo  & 13B & 69.34 & \underline{22.62} & \textbf{97.08} & \underline{63.01} \\
OmniWeaving & 8.3B & 61.78 & 22.46 & 94.12 & 59.45 \\
\midrule
\textbf{LoomVideo (Stage 3)} & 5B & 67.13 & \textbf{23.74} & 94.60 & 61.82 \\
\textbf{LoomVideo (RL)} & 5B & \textbf{70.92} & 23.59 & 94.93 & \textbf{63.15} \\

\bottomrule
        \end{tabular}
    }
\end{table*}
\textbf{Results on VBench.}
As presented in \cref{tab:quantitative-comparison-vbench}, our fully trained LoomVideo outperforms the foundational Wan 2.2 model in terms of average score and obtains the highest scores on both Imaging Quality and Overall Consistency. This demonstrates the effectiveness of our training strategy to completely replace the standard T5 text encoder with Qwen3-VL, thereby fully exploiting the rich, hierarchical multimodal alignment capabilities of the MLLM. Furthermore, it verifies that our unified architecture delivers consistent enhancements in both video generation and editing capabilities without incurring any performance degradation.

\begin{table*}[t]
    \centering
    \caption{Quantitative results on OpenVE-Bench. We re-evaluate methods with $^\dagger$ for all seven criteria.}
    \label{tab:quantitative-comparison-openve}
    \resizebox{1.0\linewidth}{!}{
        \begin{tabular}{lc|ccccccc|c}
            \toprule
\textbf{Model} & \#Params & Global Style & Background Change & Local Change & Local Remove & Local Add & Subtitle Edit & Creative Edit & \textbf{Overall} 
\\ 
\midrule
\multicolumn{10}{c}{\textit{Specialized Video Editing Models}} \\ 
\midrule
OmniVideo  & 1.3B & 1.11 & 1.18 & 1.14 & 1.14 & 1.36 & 1.00 & 2.26 & 1.31 \\
InsViE  & 2B & 2.20 & 1.06 & 1.48 & 1.36 & 1.17 & 2.18 & 2.02 & 1.64 \\
Ditto  & 14B & \textbf{4.01} & 1.68 & 2.03 & 1.53 & 1.41 & 2.81 & 1.23 & 2.10 \\
OpenVE-Edit & 5B & 3.16 & 2.36 & 2.98 & 1.85 & 2.15 & 2.91 & 2.31 & 2.53 \\ 
Kiwi-Edit$^\dagger$ & 5B & 3.62 & \underline{2.57} & \underline{3.76} & \textbf{3.36} & 2.57 & 2.91 & \underline{3.08} & \underline{3.12} \\
\midrule
\multicolumn{10}{c}{\textit{Unified Video Generation Models}} \\
\midrule
VACE  & 14B & 1.49 & 1.55 & 2.07 & 1.46 & 1.26 & 1.48 & 1.47 & 1.54 \\
VINO$^\dagger$  & 13B & \underline{3.95} & 2.39 & 3.51 & 3.20 & \underline{2.68} & 2.65 & 3.01 & 3.07 \\
UniVideo$^\dagger$  & 13B & 3.47 & \textbf{2.58} & 3.41 & 2.99 & \textbf{2.83} & 2.87 & 3.07 & 3.05 \\
OmniWeaving$^\dagger$ & 8.3B & 3.68 & 2.16 & \textbf{3.78} & 2.68 & 1.83 & \underline{3.48} & 2.8 & 2.92 \\
\midrule
\textbf{LoomVideo (Stage 2)} & 5B & 3.81 & 2.46 & 3.04 & \underline{3.33} & 2.21 & \textbf{3.64} & \textbf{3.54} & \textbf{3.15} \\
\textbf{LoomVideo (Stage 3)} & 5B & 3.62 & 2.26 & 3.32 & 2.82 & 2.40 & 3.30 & 2.86 & 2.94 \\
\textbf{LoomVideo (RL)} & 5B & 3.85 & 2.37 & 3.41 & 3.12 & 2.19 & 3.42 & 3.23 & 3.05 \\
            \bottomrule
        \end{tabular}}
\end{table*}

\textbf{Results on OpenVE-Bench.}
For instruction-based video editing, we evaluate our model on OpenVE-Bench, with results detailed in \cref{tab:quantitative-comparison-openve}. Our Stage 2 model, which primarily focuses on reconstruction and editing tasks, achieves the highest overall score. This demonstrates that our compact 5B LoomVideo can achieve state-of-the-art editing performance, underscoring its strong practicality and efficiency. Specifically, our model excels in the ``Creative Edit'' metric, indicating exceptional semantic alignment with the MLLM and a robust ability to accurately recognize and execute diverse editing intents. While our final model experiences a slight performance drop on pure instruction-based editing due to the integration of more diverse and challenging multi-modal tasks, it still remains competitive with larger-parameter baselines.

\begin{table*}[t]
\centering
\caption{Quantitative comparison on RefVIE-Bench. The \textbf{top} and the \underline{second-best} scores are only compared among open-source models.}
\label{tab:quantitative-comparison-refvie}
\resizebox{1.0\textwidth}{!}{%
\begin{tabular}{lccc|ccc|c}
\toprule
\textbf{Model} & \multicolumn{3}{c|}{\textbf{Subject Reference}} & \multicolumn{3}{c|}{\textbf{Background Reference}} & \textbf{Overall} \\
& Identity & Temporal & Physical & Reference Sim & Matting Quality & Video Quality & \\
\midrule
\multicolumn{8}{c}{\textit{Closed-Source Models}} \\
\midrule
Runway Aleph & 3.79 & 3.65 & 3.58 & 3.33 & 2.81 & 2.58 & 3.29 \\
Kling-O1 & 4.75 & 4.66 & 4.60 & 3.95 & 3.21 & 2.75 & 3.99 \\
\midrule
\multicolumn{8}{c}{\textit{Open-Source Models}} \\
\midrule
Kiwi-Edit (All data) & 3.51 & 2.96 & 2.91 & 3.40 & 2.58 & 2.40 & 2.96 \\
Kiwi-Edit (Ref. data only) & 3.98 & 3.40 & 3.34 & \underline{3.72} & \textbf{2.90} & \textbf{2.51} & 3.31 \\
VINO & 4.18 & \textbf{4.03} & \underline{3.74} & 2.93 & \underline{2.60} & 2.40 & \underline{3.53} \\
UniVideo & \underline{4.19} & 3.80 & 3.61 & 2.90 & 2.22 & 2.12 & 3.38 \\
OmniWeaving & 3.29 & 2.96 & 2.82 & 3.45 & 2.55 & 2.35 & 2.94 \\
\midrule
\textbf{LoomVideo (Stage 3)} & 4.29 & 3.90 & 2.72 & 3.75 & 2.65 & 2.38 & 3.62  \\
\textbf{LoomVideo (RL)} & \textbf{4.50} & \underline{3.98} & \textbf{3.90} & \textbf{3.88} & \textbf{2.90} & \underline{2.50} & \textbf{3.78} \\
\bottomrule
\end{tabular}
}
\end{table*}
\textbf{Results on RefVIE-Bench.}
We further evaluate our model's performance on instruction-and-reference-guided video editing (incorporating a single subject or background reference image) using RefVIE-Bench. As shown in \cref{tab:quantitative-comparison-refvie}, LoomVideo achieves the highest overall score among all open-source baseline methods, outperforming the second-place VINO by a significant margin of 7\%. This substantial improvement highlights our model's robust capability and precision in handling fine-grained visual conditions.

\begin{table*}[t]
    \centering
    \caption{Quantitative comparison of the TIV2V task in IntelligentVBench.}
    \label{tab:quantitative-comparison-intelligentvbench-tiv2v}
    \resizebox{0.5\linewidth}{!}{
        \begin{tabular}{lc|cccc}
\toprule
\multirow{2}{*}{Model}    & \multirow{2}{*}{\#Params} & \multicolumn{4}{c}{\textbf{TIV2V}}  \\
&  & $\mathcal{IF}\uparrow$ & $\mathcal{CP}\uparrow$ & $\mathcal{VQ}\uparrow$ & $\mathbf{AVG}$ \\
\midrule
VACE-Wan2.1  &  14B & 1.46 & 1.42 & 1.71 & 1.53 \\
VACE-LTX  & 2B & 1.43 & 1.36 & 1.25 & 1.35 \\
VINO  & 13B & 2.86 & 2.90 & 2.52 & 2.76 \\
UniVideo (query)  & 13B & 3.22 & 3.91 & 3.26 & 3.46 \\
UniVideo (hidden)  & 13B & 3.13 & 4.01 & 2.93 & 3.36 \\
OmniWeaving & 8.3B & \underline{4.00} & \underline{4.04} & \underline{3.65} & \underline{3.89} \\
\midrule
\textbf{LoomVideo (Stage 3)} & 5B & 4.35 & 4.08 & 3.99 & 4.14 \\
\textbf{LoomVideo (RL)} & 5B & \textbf{4.39} & \textbf{4.23} & \textbf{4.08} & \textbf{4.24} \\
\bottomrule
        \end{tabular}}
\end{table*}
\begin{table*}[t]
    \centering
    \caption{Quantitative comparison of the Compositional MI2V task in IntelligentVBench.}
    \label{tab:quantitative-comparison-intelligentvbench-mi2v}
    \resizebox{1.0\linewidth}{!}{
        \begin{tabular}{lc|cccc|cccc|cccc}
            \toprule
\multirow{2}{*}{Model}    & \multirow{2}{*}{\#Params}&\multicolumn{4}{c}{\textbf{1Subject (with BKG)}}  &
        \multicolumn{4}{c}{\textbf{2Subjects (with BKG)}}  &
        \multicolumn{4}{c}{\textbf{3Subjects (with BKG)}}  \\
&  & $\mathcal{IF}\uparrow$ & $\mathcal{CP}\uparrow$ & $\mathcal{VQ}\uparrow$ & $\mathbf{AVG}$ & $\mathcal{IF}\uparrow$ & $\mathcal{CP}\uparrow$ & $\mathcal{VQ}\uparrow$ & $\mathbf{AVG}$ & $\mathcal{IF}\uparrow$ & $\mathcal{CP}\uparrow$ & $\mathcal{VQ}\uparrow$ & $\mathbf{AVG}$ \\
\midrule
\multicolumn{13}{c}{\textit{Specialized Video Generation Models}} \\ 
\midrule
SkyReels-A2  & 14B & 3.51 & 4.08 & 4.46 & 4.02 & 3.22 & 3.76 & 4.37 & 3.78 & 1.64 & 1.76 & 2.50 & 1.97 \\
SkyReels-V3  & 14B & 3.46 & 3.71 & \underline{4.65} & 3.94 & 3.28 & 3.84 & 4.44 & 3.86 & 2.59 & \underline{3.10} & 4.30 & 3.33 \\
Phantom  & 14B & 3.21 & 2.95 & 4.29 & 3.48 & 2.88 & 3.42 & 4.38 & 3.55 & 2.36 & 2.79 & 4.21 & 3.12 \\
\midrule
\multicolumn{13}{c}{\textit{Unified Video Generation Models}} \\
\midrule
VACE-Wan2.1  & 14B & 3.88 & \underline{4.48} & \textbf{4.68} & \underline{4.35} & 3.31 & 4.03 & 4.51 & 3.95 & 2.60 & 3.03 & \underline{4.40} & \underline{3.34} \\
VACE-LTX  & 2B & 2.74 & 2.86 & 2.89 & 2.83 & 2.12 & 2.26 & 2.49 & 2.29 & 1.94 & 2.06 & 2.41 & 2.14 \\
VINO  & 13B & 3.72 & 4.22 & 4.46 & 4.13 & 3.56 & \textbf{4.34} & \textbf{4.58} & \underline{4.16} & \underline{2.63} & 2.97 & 4.24 & 3.28 \\
UniVideo(query)  & 13B & 3.35 & 3.90 & 4.41 & 3.89 & 2.98 & 3.73 & 4.18 & 3.63 & 2.30 & 2.50 & 3.89 & 2.89 \\ 
UniVideo(hidden)  & 13B & 3.33 & 4.18 & 4.38 & 3.97 & 3.22 & 4.12 & 4.36 & 3.90 & 2.31 & 2.83 & 3.94 & 3.03 \\
OmniWeaving & 8.3B & \textbf{4.35} & \textbf{4.53} & 4.58 & \textbf{4.49} & \textbf{4.08} & \underline{4.22} & \underline{4.52} & \textbf{4.27} & \textbf{3.53} & \textbf{4.01} & \textbf{4.54} & \textbf{4.03} \\
\midrule
\textbf{LoomVideo (Stage 3)} & 5B & 3.81 & 3.78 & 3.91 & 3.83 & 3.30 & 3.32 & 3.69 & 3.44 & 2.40 & 2.70 & 3.61 & 2.90 \\
\textbf{LoomVideo (RL)} & 5B & \underline{3.95} & 3.90 & 4.20 & 4.02 & \underline{3.51} & 3.44 & 4.06 & 3.67 & 2.47 & 2.67 & 3.67 & 2.94 \\
         
            \bottomrule
        \end{tabular}}
\end{table*}
\textbf{Results on IntelligentVBench.}
To assess our model's multimodal conditioning capabilities, we report results on the Text-Image-to-Video (TIV2V) and Multi-Image-to-Video (MI2V) sub-tasks of IntelligentVBench, detailed in \cref{tab:quantitative-comparison-intelligentvbench-tiv2v} and \cref{tab:quantitative-comparison-intelligentvbench-mi2v}. The TIV2V task shares inherent similarities with RefVIE, and our 5B model consistently secures the top position, outperforming the second-best OmniWeaving by 8\%. For the highly complex compositional MI2V task, LoomVideo achieves performance comparable to the SOTA UniVideo (query) model. We attribute this competitive but non-leading performance primarily to the inherent capacity limits of our 5B-parameter scale, as well as to domain shift, given that our MI2V training data predominantly consists of Taobao e-commerce scenarios rather than the open-domain distribution of IntelligentVBench.

\begin{table*}[t]
\centering
\caption{Quantitative comparison on FashionVideoBench.}
\label{tab:quantitative-comparison-fashionvideobench}
\resizebox{1.0\textwidth}{!}{%
\begin{tabular}{lccc|cccccc|c}
\toprule
\multirow{2}{*}{\textbf{Model}} & \multicolumn{3}{c|}{\textbf{Split by Metrics}} & \multicolumn{6}{c|}{\textbf{Split by Task}} & \multirow{2}{*}{\textbf{Overall}} \\
& SC & PF & VQ & Product Edit & Model Edit & Freeform Edit & PRef Edit & MRef Edit & MI2V & \\
\midrule
UniVideo & 4.08 & 4.34 & 4.37 & 3.84 & \underline{4.93} & 4.20 & 4.24 & 4.05 & 4.29 & 4.26 \\
OmniWeaving & 3.28 & 3.71 & 3.70 & 3.67 & 4.04 & 3.49 & 2.95 & 3.49 & 3.72 & 3.56 \\
VINO & \underline{4.18} & \underline{4.51} & \underline{4.45} & \underline{4.02} & 4.83 & \underline{4.27} & \underline{4.27} & \textbf{4.43} & \underline{4.45} & 4.38 \\
\midrule
\textbf{LoomVideo (Stage 3)} & \textbf{4.45} & \textbf{4.74} & 4.61 & \textbf{4.59} & \textbf{4.95} & \textbf{4.47} & \textbf{4.51} & \underline{4.37} & \textbf{4.70} & \textbf{4.60} \\
\textbf{LoomVideo (RL)} & 4.44 & 4.71 & \textbf{4.62} & 4.59 & 4.92 & 4.45 & 4.51 & 4.37 &4.70 & 4.59 \\
\bottomrule
\end{tabular}
}
\end{table*}

\textbf{Results on FashionVideoBench.}
Recognizing that existing methods primarily focus on general domains and lack specialized evaluations for e-commerce applications, we introduce FashionVideoBench, constructed entirely from held-out internal data. As demonstrated in~\cref{tab:quantitative-comparison-fashionvideobench}, LoomVideo consistently achieves the highest overall scores across all six core sub-tasks compared to open-source counterparts. This comprehensive superiority strongly validates our model's exceptional practicality, controllability, and generation quality in complex e-commerce and fashion generation scenarios.

\textbf{Ablation Study on Reinforcement Learning Post-Training.} 
To validate the necessity of RL post-training, we conduct comprehensive comparisons between our Stage 3 model and the final model across all evaluation benchmarks. As evidenced by the quantitative results presented in all aforementioned tables, our final model achieves significant performance improvements on all benchmarks except FashionVideoBench, where our model has already attained a high level of editing quality with limited room for further enhancement. This result strongly validates that RL post-training effectively boosts the model's video generation and editing capabilities, serving as a critical component for our training pipeline.

\subsection{Qualitative Comparison}
\begin{figure}[!p]
    \centering
    \includegraphics[width=\textwidth]{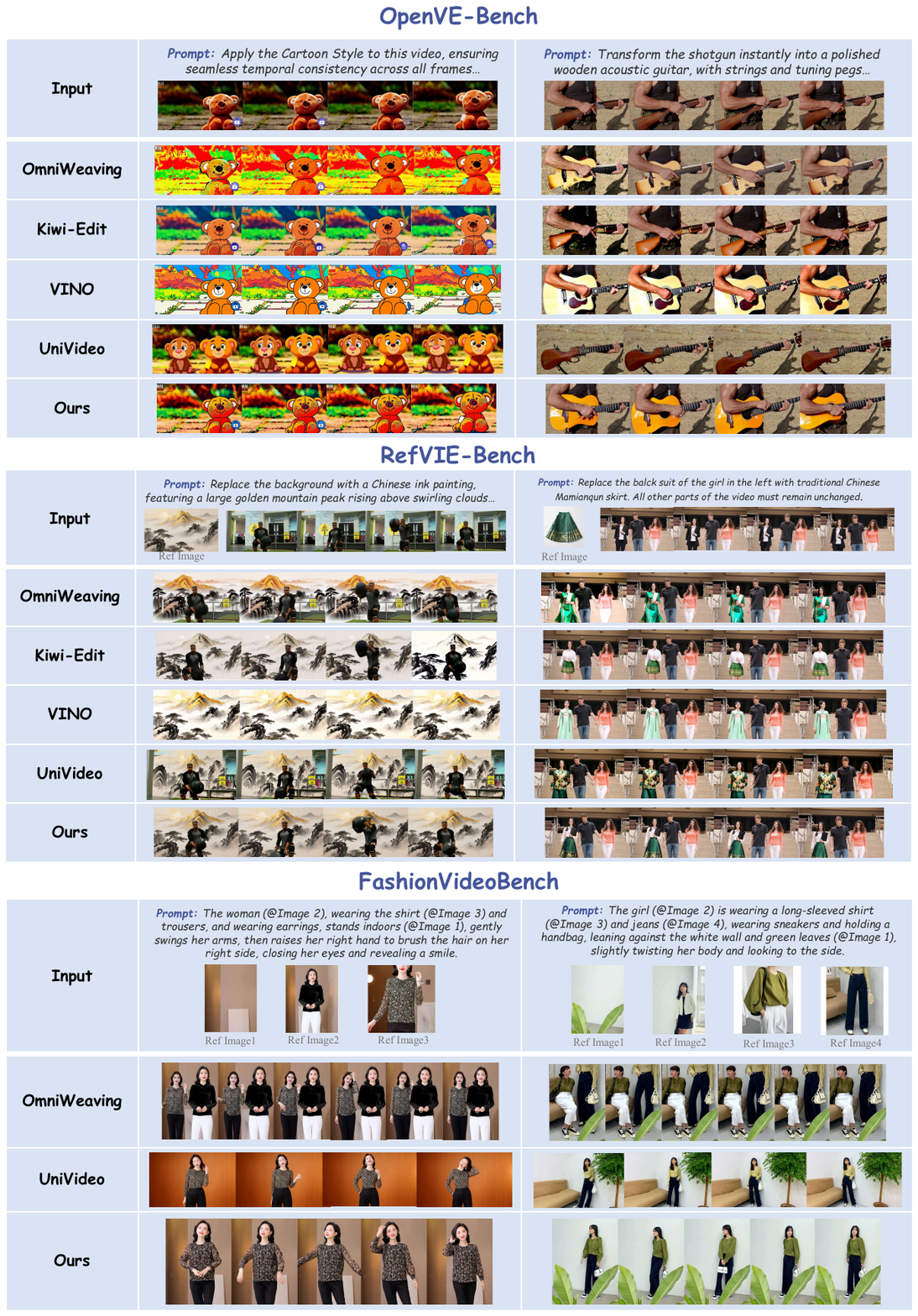} 
    \vspace{-1em}
    \caption{Qualitative comparison of LoomVideo and other baselines.}
    \vspace{-2em}
    \label{fig:comparison}
\end{figure}

As illustrated in~\cref{fig:comparison}, we qualitatively compare four methods, including Kiwi-Edit, VINO, UniVideo and OmniWeaving. In the traditional video editing tasks on OpenVE-Bench, our model exhibits superior instruction following capability and editing quality compared with all baseline methods. For the reference-guided editing tasks on RefVIE-Bench, our model can more accurately restore reference images and produce high-fidelity editing results. In the MI2V tasks on FashionVideoBench, our model achieves precise restoration of each reference image while maintaining favorable overall video generation quality. More qualitative results of our method are provided in ~\cref{fig:case1}, ~\cref{fig:case2}, ~\cref{fig:case3}, and ~\cref{fig:case4} in Appendix~\ref{app:2}.

\begin{table*}[t]
    \centering
    \caption{\textbf{Comparison of inference time.} We test all the models on the same type of GPU. We measure the average time required to generate/edit a single 480$\times$832$\times$97 video for both Text-to-Video (T2V) and Instruction-based Video Editing (TV2V) tasks. The speedup ratio is calculated based on OmniWeaving.}
    \label{tab:comparison-inference-time}
    \resizebox{0.8\textwidth}{!}{
        \begin{tabular}{lccccc}
            \toprule
            \textbf{Model} & \textbf{\#Params} & \textbf{Source Token Injection} & \textbf{T2V (s)} & \textbf{TV2V (s)} \\
            \midrule
            Wan 2.2           & 5B + 5.68B (UMT5-XXL)  & -  & 138.61    & -      \\
            \midrule
            UniVideo (hidden) & 13B+ 7B    & Token Concat   & 1792.65   & 6140.18 \\
            OmniWeaving       & 8.3B+ 7B   & Channel Concat   & 824.93    & 899.32 \\
            VINO              & 13B+ 4B    & Token Concat   & 2793.52   & 9555.13 \\
            \midrule
            Ours              & 5B + 8B    & Scale-and-Add  & \textbf{132.23 (6.24$\times$)}    &  \textbf{166.30 (5.41$\times$)}     \\
            \bottomrule
        \end{tabular}
    }
\end{table*}
\subsection{Efficiency Comparison}

As shown in~\cref{tab:comparison-inference-time}, our model achieves significantly faster inference speeds compared to counterpart models across both T2V generation and video editing tasks. Compared to the fastest baseline, OmniWeaving, LoomVideo demonstrates a remarkable speedup of 6.24$\times$ for T2V generation and 5.41$\times$ for video editing. Furthermore, for models such as UniVideo and VINO that rely on the token concatenation strategy, the inference time required to edit a single video is approximately four times their respective T2V generation time. This observation aligns with our analysis that token concatenation inherently quadruples the computational complexity of the self-attention mechanism. Overall, these results compellingly highlight the exceptional efficiency and practical viability of our proposed architecture in this domain.

\section{Conclusion and Future Work}

In this work, we introduce LoomVideo, a highly efficient unified framework capable of handling diverse video generation and editing tasks via interleaved multimodal inputs. By replacing the standard text encoder with an MLLM, we unlock rich, hierarchical semantic understanding. To fully leverage this, we proposed the \textbf{Deepstack} injection mechanism, ensuring deep alignment between multimodal controls and the spatiotemporal generative prior. Furthermore, we address the severe computational bottleneck of existing editing frameworks by introducing the zero-overhead \textbf{Scale-and-Add} conditioning approach. By fundamentally bypassing the conventional token concatenation paradigm, this elegant design enables LoomVideo to achieve over 5.4$\times$ inference speedup while maintaining robust capabilities for complex, non-rigid edits. Coupled with the \textbf{Negative Temporal RoPE} strategy for multi-reference guidance, extensive experiments demonstrate that our compact 5B-parameter model achieves state-of-the-art or highly competitive performance across comprehensive open-domain benchmarks. Notably, it exhibits exceptional superiority and practicality in complex e-commerce and fashion generation scenarios.

Several avenues remain for future exploration. We aim to scale up the diffusion transformer's parameter size and extend our multi-resolution training pipeline to support higher-definition (e.g., 720p or 1080p) and longer-duration video generation, ultimately pushing the boundaries of physical world simulation and visual fidelity.
\section{Contributions and Acknowledgments}

Authors are organized by their affiliation. 
$^*$ indicates equal contribution, $^\S$ indicates corresponding authors, and $^\dagger$ indicates the project leader.

\medskip
\noindent
\textbf{Peking University:} 
Jianzong Wu$^{*\dagger}$, Hao Lian$^*$, Jiongfan Yang, Dachao Hao, Ye Tian, Yunhai Tong$^\S$

\noindent
\textbf{Alibaba Group:} 
Jingyuan Zhu, Biaolong Chen, Qiaosong Qi, Aixi Zhang, Wanggui He, Mushui Liu, Jinlong Liu, Pipei Huang, Hao Jiang$^\S$

\medskip
We would like to thank MSALab at Peking University and Alibaba Group for their support and discussions throughout this project.

\bibliography{main}
\bibliographystyle{iclr2025_conference}

\clearpage

\appendix

\section*{Appendix}

\section{More Experimental Results}
\label{app:2}

More qualitative results across various tasks are shown in~\cref{fig:case1}, ~\cref{fig:case2}, ~\cref{fig:case3}, and ~\cref{fig:case4}. The results demonstrate that LoomVideo unifies diverse video generation capabilities within a single model. Notably, it achieves this versatility while maintaining high efficiency, powered by a compact 5B parameter architecture.

\section{Limitation}

\begin{figure}[h]
    \centering
    \includegraphics[width=\textwidth]{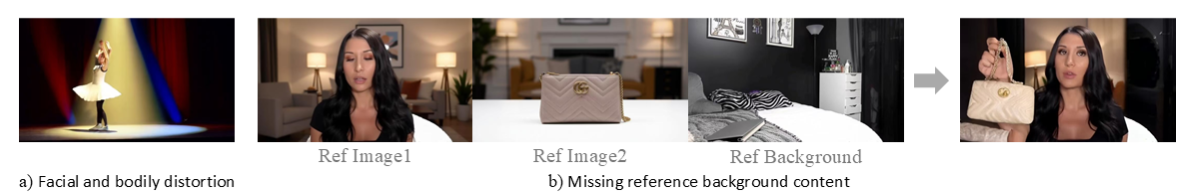} 
    \vspace{-1em}
    \caption{Low-quality generation cases of LoomVideo.}
    \vspace{-2em}
    \label{fig:limit}
\end{figure}

\begin{figure}[t]
    \centering
    \includegraphics[width=\textwidth]{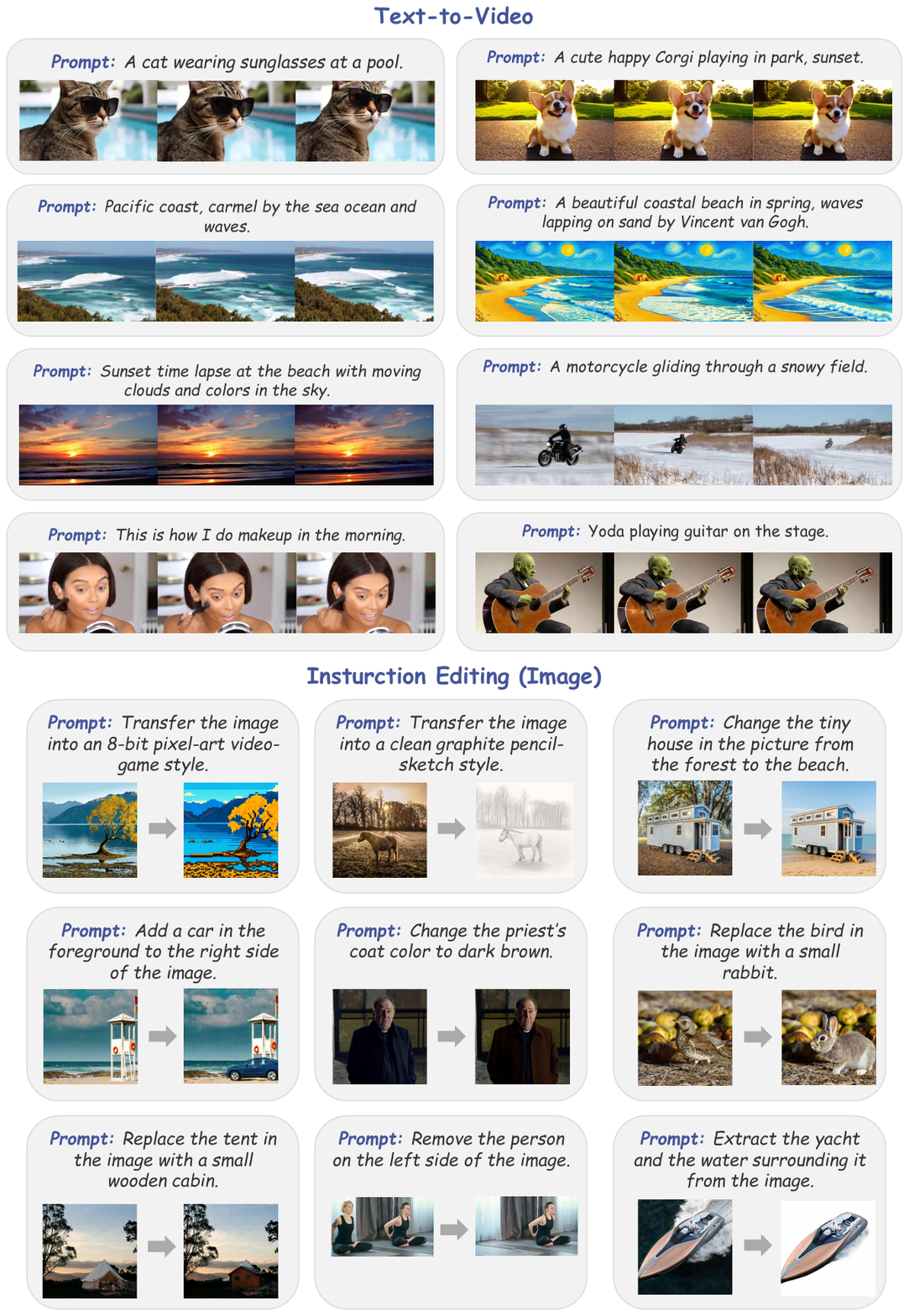} 
    \vspace{-1em}
    \caption{Qualitative results for LoomVideo on Text-to-Video and Instuction Editing tasks.}
    \vspace{-2em}
    \label{fig:case1}
\end{figure}

\begin{figure}[t]
    \centering
    \includegraphics[width=\textwidth]{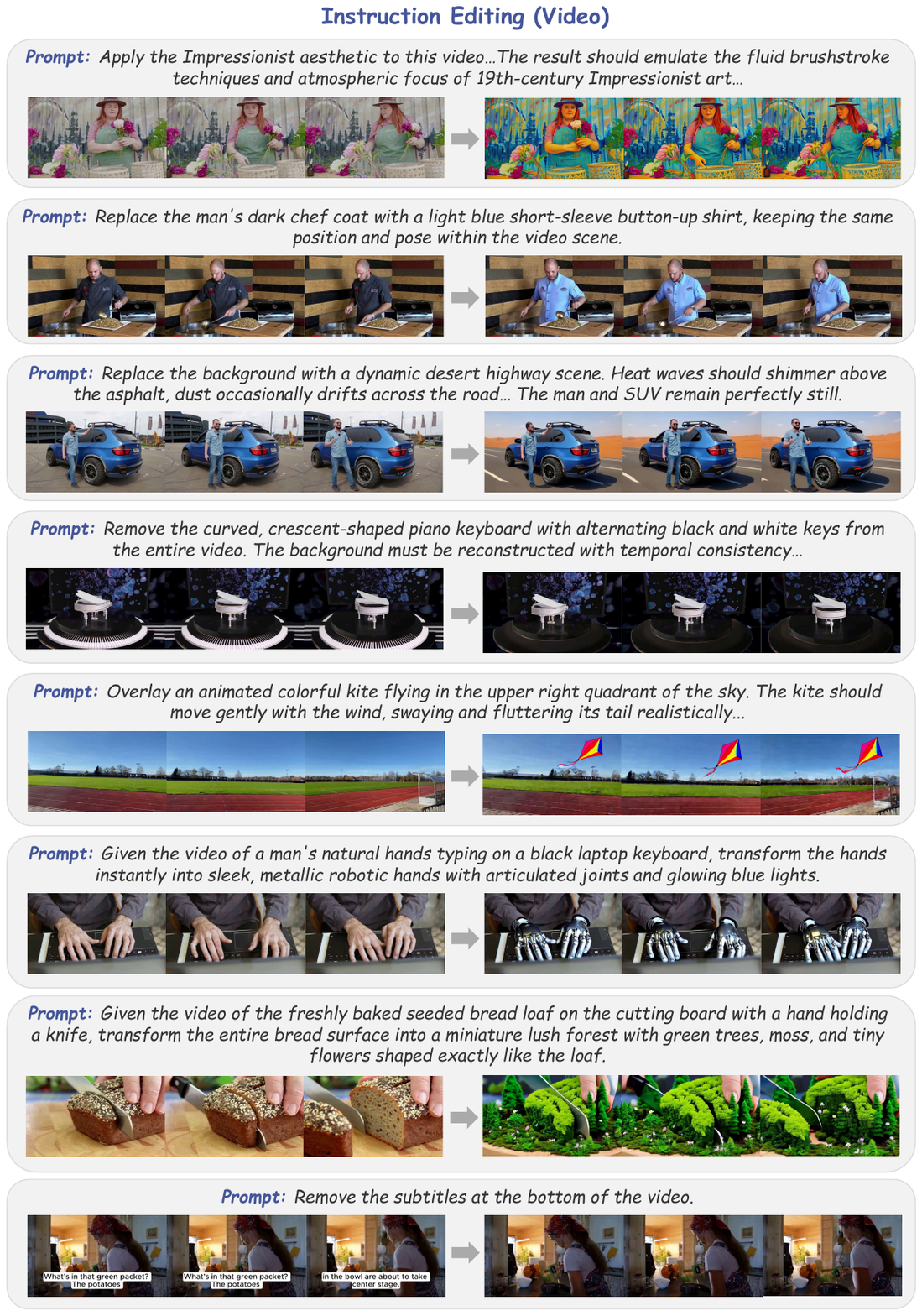} 
    \vspace{-1em}
    \caption{Qualitative results for LoomVideo on Instuction Editing task.}
    \vspace{-2em}
    \label{fig:case2}
\end{figure}

\begin{figure}[t]
    \centering
    \includegraphics[width=\textwidth]{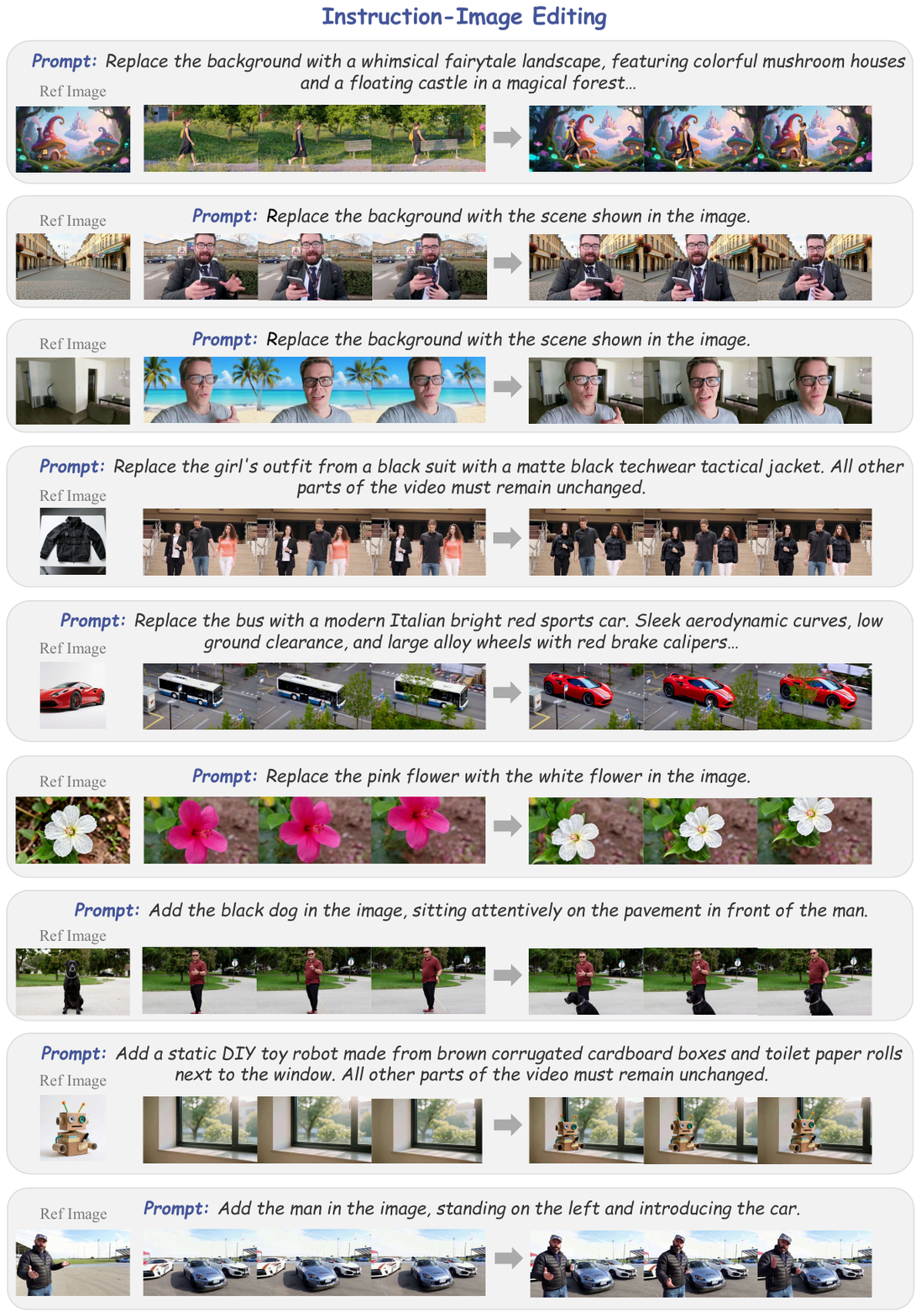} 
    \vspace{-1em}
    \caption{Qualitative results for LoomVideo on Instuction-Image Editing task.}
    \vspace{-2em}
    \label{fig:case3}
\end{figure}

\begin{figure}[t]
    \centering
    \includegraphics[width=\textwidth]{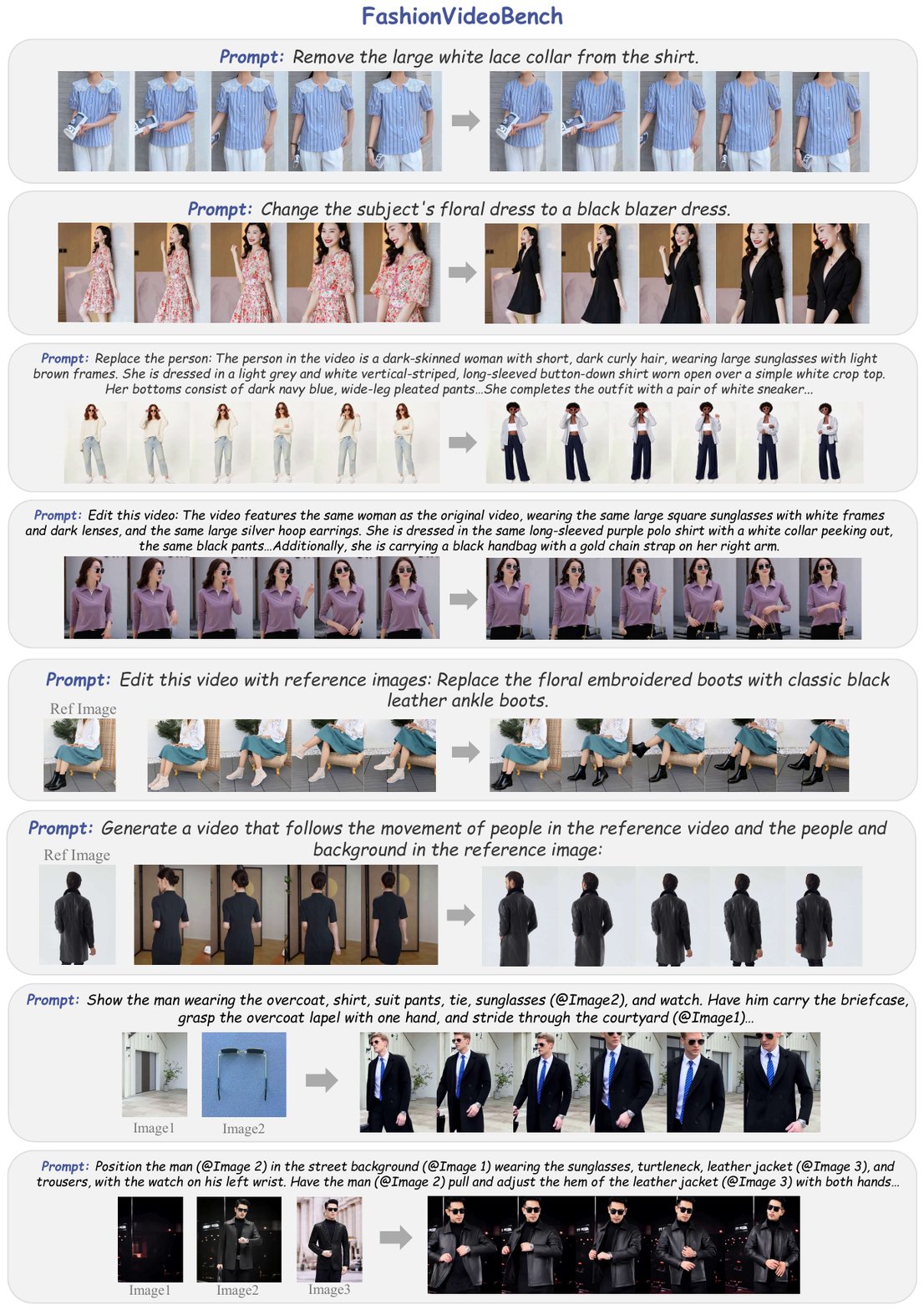} 
    \vspace{-1em}
    \caption{Qualitative results for LoomVideo on our benchmark FashionVideoBench.}
    \vspace{-2em}
    \label{fig:case4}
\end{figure}

As shown in \cref{fig:limit}, despite achieving competitive performance across various video generation tasks, our model still presents several limitations for future exploration. First, noticeable visual artifacts including distorted eyes and unnatural limb movements emerge under highly dynamic scenarios. This problem stems from our relatively compact model size (5B parameters), which constrains the model’s capability to learn fine-grained temporal dynamics. Second, our training data is primarily curated for e-commerce scenarios and lacks sufficient diversity across real-world scenes. This constraint hinders accurate background reconstruction on general MI2V benchmarks such as IntelligentVBench. In future work, we plan to enhance model capacity and expand the coverage of training data to address the above issues.

\section{FashionVideoBench Evaluation Prompts}
\label{app:3}

This section presents the evaluation prompt templates for each task in FashionVideoBench. 
Specifically, the evaluation prompt template for product editing is detailed in~\cref{fig:bench_product_edit}.
The evaluation prompt template for model editing is detailed in~\cref{fig:bench_model_edit}.
The evaluation prompt template for freeform editing is detailed in~\cref{fig:bench_freeform_edit}.
Finally, we design evaluation prompt templates for the distinct sub-tasks of reference-guided video editing in ~\cref{fig:bench_item_replace_withref}, ~\cref{fig:bench_motion_transfer_withref}, and ~\cref{fig:bench_multiref}.

\begin{figure}[ht]

\begin{AcademicBox}[Evaluation Prompt Template for Product Editing]
You are an expert data rater specializing in grading video product editing. You will be provided with an original video, the edited video, and the editing instruction. The instruction specifies changes to be made to a product in the video. \\
Your task is to evaluate the editing performance on a 5-point scale across three key dimensions: \\
\textbf{1. The first score: Subject Consistency} \\
\textbf{Objective:} Evaluate how well the edited product maintains its identity and appearance throughout the video. \\
\textbf{- 5:} Perfect Preservation. Product identity is flawlessly maintained with all attributes correctly edited.
\textbf{- 4:} High Preservation. Product is highly consistent with only minor detail loss.
\textbf{- 3:} Moderate Preservation. Product is generally recognizable but with noticeable flaws.
\textbf{- 2:} Low Preservation. Product suffers from severe drift or incorrect edits.
\textbf{- 1:} Complete Failure. Product is completely irrelevant or incorrectly edited. \\
\textbf{2. The second score: Prompt Following} \\
\textbf{Objective:} Assess how accurately the video executes the editing instruction. \\
\textbf{- 5:} Perfect Alignment. Every detail of the instruction is perfectly rendered.
\textbf{- 4:} Good Alignment. The core semantics of the prompt are successfully captured.
\textbf{- 3:} Partial Alignment. The model captures the main idea but ignores specific nuances.
\textbf{- 2:} Weak Alignment. The video mostly ignores the prompt.
\textbf{- 1:} No Alignment. The video content has no relevance to the instruction. \\
\textbf{3. The third score: Visual Quality} \\
\textbf{Objective:} Evaluate the general aesthetic quality, temporal consistency, and motion smoothness. \\
\textbf{- 5:} Excellent. Exceptional visual aesthetics and perfect temporal consistency.
\textbf{- 4:} Good. High overall quality with only minor artifacts.
\textbf{- 3:} Fair. Acceptable but noticeably flawed.
\textbf{- 2:} Poor. Significantly degraded quality.
\textbf{- 1:} Unacceptable. Completely collapsed visual integrity. \\

\textbf{Example Response Format:}
You are required to return a dictionary structured as follows: \{"Subject Consistency": [A number from 1 to 5], "Prompt Following": [A number from 1 to 5], "Visual Quality": [A number from 1 to 5]\}. \\

The editing instruction is: \textless edit\_prompt\textgreater \\
Below are the original video and edited video:

\end{AcademicBox}
\vspace{-1em}
\caption{Evaluation prompt template for video product editing.\looseness=-1}
\label{fig:bench_product_edit}
\end{figure}
\begin{figure}[t]

\begin{AcademicBox}[Evaluation Prompt Template for Video Model Replacement Editing]
You are an expert data rater specializing in grading video model replacement editing. You will be provided with an original video, the edited video, and the editing instruction. The instruction specifies replacing a model or person in the video. \\
Your task is to evaluate the editing performance on a 5-point scale across three key dimensions: \\
\textbf{1. The first score: Subject Consistency} \\
\textbf{Objective:} Evaluate how well the new model maintains consistency throughout the video. \\
\textbf{- 5:} Perfect Consistency. Model identity is flawlessly maintained across all frames.
\textbf{- 4:} High Consistency. Model is highly consistent with only minor flickering.
\textbf{- 3:} Moderate Consistency. Model is recognizable but shows some instability.
\textbf{- 2:} Low Consistency. Model suffers from significant identity drift.
\textbf{- 1:} Complete Failure. Model is completely inconsistent or unrecognizable. \\
\textbf{2. The second score: Prompt Following} \\
\textbf{Objective:} Assess how accurately the video executes the model replacement instruction. \\
\textbf{- 5:} Perfect Alignment. The model is perfectly replaced as specified.
\textbf{- 4:} Good Alignment. The model is correctly replaced with minor deviations.
\textbf{- 3:} Partial Alignment. The model is partially replaced but with noticeable errors.
\textbf{- 2:} Weak Alignment. The replacement is attempted but poorly executed.
\textbf{- 1:} No Alignment. The model was not replaced or replaced incorrectly. \\
\textbf{3. The third score: Visual Quality} \\
\textbf{Objective:} Evaluate the general aesthetic quality, temporal consistency, and motion smoothness. \\
\textbf{- 5:} Excellent. Exceptional visual aesthetics and perfect temporal consistency.
\textbf{- 4:} Good. High overall quality with only minor artifacts.
\textbf{- 3:} Fair. Acceptable but noticeably flawed.
\textbf{- 2:} Poor. Significantly degraded quality.
\textbf{- 1:} Unacceptable. Completely collapsed visual integrity. \\

\textbf{Example Response Format:}
You are required to return a dictionary structured as follows: \{"Subject Consistency": [A number from 1 to 5], "Prompt Following": [A number from 1 to 5], "Visual Quality": [A number from 1 to 5]\}. \\

The editing instruction is: \textless edit\_prompt\textgreater \\
Below are the original video and edited video:

\end{AcademicBox}
\vspace{-1em}
\caption{Evaluation prompt template for video model replacement editing.\looseness=-1}
\label{fig:bench_model_edit}
\end{figure}
\begin{figure}[t]

\begin{AcademicBox}[Evaluation Prompt Template for Freeform Video Editing]
You are an expert data rater specializing in grading freeform video editing. You will be provided with an original video, the edited video, and the editing instruction. The instruction specifies general editing changes to be made to the video. \\
Your task is to evaluate the editing performance on a 5-point scale across three key dimensions: \\
\textbf{1. The first score: Subject Consistency} \\
\textbf{Objective:} Evaluate how well the edited content maintains consistency throughout the video. \\
\textbf{- 5:} Perfect Consistency. Edited content is flawlessly maintained across all frames.
\textbf{- 4:} High Consistency. Content is highly consistent with only minor flickering.
\textbf{- 3:} Moderate Consistency. Content is recognizable but shows some instability.
\textbf{- 2:} Low Consistency. Content suffers from significant identity drift.
\textbf{- 1:} Complete Failure. Content is completely inconsistent or unrecognizable. \\
\textbf{2. The second score: Prompt Following} \\
\textbf{Objective:} Assess how accurately the video executes the editing instruction. \\
\textbf{- 5:} Perfect Alignment. Every detail of the instruction is perfectly rendered.
\textbf{- 4:} Good Alignment. The core semantics of the prompt are successfully captured.
\textbf{- 3:} Partial Alignment. The model captures the main idea but ignores specific nuances.
\textbf{- 2:} Weak Alignment. The video mostly ignores the prompt.
\textbf{- 1:} No Alignment. The video content has no relevance to the instruction. \\
\textbf{3. The third score: Visual Quality} \\
\textbf{Objective:} Evaluate the general aesthetic quality, temporal consistency, and motion smoothness. \\
\textbf{- 5:} Excellent. Exceptional visual aesthetics and perfect temporal consistency.
\textbf{- 4:} Good. High overall quality with only minor artifacts.
\textbf{- 3:} Fair. Acceptable but noticeably flawed.
\textbf{- 2:} Poor. Significantly degraded quality.
\textbf{- 1:} Unacceptable. Completely collapsed visual integrity. \\

\textbf{Example Response Format:}
You are required to return a dictionary structured as follows: \{"Subject Consistency": [A number from 1 to 5], "Prompt Following": [A number from 1 to 5], "Visual Quality": [A number from 1 to 5]\}. \\

The editing instruction is: \textless edit\_prompt\textgreater \\
Below are the original video and edited video:

\end{AcademicBox}
\vspace{-1em}
\caption{Evaluation prompt template for freeform video editing.\looseness=-1}
\label{fig:bench_freeform_edit}
\end{figure}
\begin{figure}[t]

\begin{AcademicBox}[Evaluation Prompt Template for Video Item Replacement with Reference Image]
You are an expert data rater specializing in grading video item replacement with reference image. You will be provided with an original video, a reference image, the edited video, and the editing instruction. The instruction specifies replacing an item in the video using the reference image. \\
Your task is to evaluate the editing performance on a 5-point scale across three key dimensions: \\
\textbf{1. The first score: Subject Consistency} \\
\textbf{Objective:} Evaluate how well the replaced item matches the reference image and maintains consistency throughout the video. \\
\textbf{- 5:} Perfect Match. The replaced item is identical to the reference image and consistent across all frames.
\textbf{- 4:} High Match. The replaced item closely matches the reference with minor differences.
\textbf{- 3:} Moderate Match. The replaced item is similar but has noticeable differences.
\textbf{- 2:} Low Match. The replaced item has significant differences from the reference.
\textbf{- 1:} No Match. The replaced item is completely different from the reference. \\
\textbf{2. The second score: Prompt Following} \\
\textbf{Objective:} Assess how accurately the video executes the item replacement instruction. \\
\textbf{- 5:} Perfect Alignment. The item is perfectly replaced as specified.
\textbf{- 4:} Good Alignment. The item is correctly replaced with minor deviations.
\textbf{- 3:} Partial Alignment. The item is partially replaced but with noticeable errors.
\textbf{- 2:} Weak Alignment. The replacement is attempted but poorly executed.
\textbf{- 1:} No Alignment. The item was not replaced or replaced incorrectly. \\
\textbf{3. The third score: Visual Quality} \\
\textbf{Objective:} Evaluate the general aesthetic quality, temporal consistency, and motion smoothness. \\
\textbf{- 5:} Excellent. Exceptional visual aesthetics and perfect temporal consistency.
\textbf{- 4:} Good. High overall quality with only minor artifacts.
\textbf{- 3:} Fair. Acceptable but noticeably flawed.
\textbf{- 2:} Poor. Significantly degraded quality.
\textbf{- 1:} Unacceptable. Completely collapsed visual integrity. \\

\textbf{Example Response Format:}
You are required to return a dictionary structured as follows: \{"Subject Consistency": [A number from 1 to 5], "Prompt Following": [A number from 1 to 5], "Visual Quality": [A number from 1 to 5]\}. \\

The editing instruction is: \textless edit\_prompt\textgreater \\
Below are the original video, reference image, and edited video:

\end{AcademicBox}
\vspace{-1em}
\caption{Evaluation prompt template for video item replacement with reference image.\looseness=-1}
\label{fig:bench_item_replace_withref}
\end{figure}
\begin{figure}[t]

\begin{AcademicBox}[Evaluation Prompt Template for Motion-Transfer Video Generation]
You are an expert data rater specializing in grading motion-transfer video generation. You will be provided with: (1) A source video (providing the motion/pose reference), (2) A reference image (providing the target person appearance and background), and (3) The generated video (which should combine the source motion with the reference appearance). \\
Your task is to evaluate the generated video on a 5-point scale across three key dimensions: \\
\textbf{1. The first score: Subject Consistency} \\
\textbf{Objective:} Evaluate both (a) how faithfully the generated video reproduces the person's appearance (face, body, clothing) and background from the reference image, and (b) how consistently these elements are maintained across all frames of the generated video. \\
\textbf{- 5:} Perfect. Reference appearance is faithfully reproduced and flawlessly consistent across all frames.
\textbf{- 4:} High Consistency. Appearance closely matches the reference with only minor detail loss or flickering.
\textbf{- 3:} Moderate. Person is generally recognizable from the reference but with noticeable differences or temporal instability.
\textbf{- 2:} Low Consistency. Significant identity drift from the reference or severe frame-to-frame inconsistency.
\textbf{- 1:} Complete Failure. Generated appearance is unrelated to the reference or completely inconsistent. \\
\textbf{2. The second score: Prompt Following} \\
\textbf{Objective:} Assess how faithfully the generated video reproduces the movements, poses, and motion trajectory from the source video as specified by the instruction. \\
\textbf{- 5:} Perfect Alignment. All movements, poses, and temporal dynamics from the source are flawlessly reproduced.
\textbf{- 4:} Good Alignment. Core motion is accurately captured with only minor timing or pose deviations.
\textbf{- 3:} Partial Alignment. General movement pattern is recognizable but with noticeable errors in pose or timing.
\textbf{- 2:} Weak Alignment. Motion is largely distorted or only partially follows the source.
\textbf{- 1:} No Alignment. Generated motion bears no resemblance to the source video. \\
\textbf{3. The third score: Visual Quality} \\
\textbf{Objective:} Evaluate the general aesthetic quality, temporal consistency, and motion smoothness. \\
\textbf{- 5:} Excellent. Exceptional visual aesthetics and perfect temporal consistency.
\textbf{- 4:} Good. High overall quality with only minor artifacts.
\textbf{- 3:} Fair. Acceptable but noticeably flawed.
\textbf{- 2:} Poor. Significantly degraded quality.
\textbf{- 1:} Unacceptable. Completely collapsed visual integrity. \\

\textbf{Example Response Format:}
You are required to return a dictionary structured as follows: \{"Subject Consistency": [A number from 1 to 5], "Prompt Following": [A number from 1 to 5], "Visual Quality": [A number from 1 to 5]\}. \\

The editing instruction is: \textless edit\_prompt\textgreater \\
Below are the source video, reference image, and generated video:

\end{AcademicBox}
\vspace{-1em}
\caption{Evaluation prompt template for motion-transfer video generation.\looseness=-1}
\label{fig:bench_motion_transfer_withref}
\end{figure}
\begin{figure}[t]

\begin{AcademicBox}[Evaluation Prompt Template for Multi-Reference-Image-to-Video Generation]
You are an expert data rater specializing in grading multi-reference-image-to-video generation. You will be provided with multiple reference images and the generated video. The reference images provide visual elements (person appearance, clothing items, background, etc.) that should appear in the generated video. A text instruction describes how to compose these elements into a coherent video. \\
Your task is to evaluate the generated video on a 5-point scale across three key dimensions: \\
\textbf{1. The first score: Subject Consistency} \\
\textbf{Objective:} Evaluate both (a) how faithfully the generated video reproduces the visual elements from the reference images (person identity, clothing details, background, product appearance, etc.), and (b) how consistently these elements are maintained across all frames of the generated video. \\
\textbf{- 5:} Perfect. All referenced elements are faithfully reproduced and flawlessly consistent across all frames.
\textbf{- 4:} High Consistency. Most elements accurately match the references with only minor detail loss or flickering.
\textbf{- 3:} Moderate. Elements are generally recognizable but with noticeable differences from references or temporal instability.
\textbf{- 2:} Low Consistency. Significant mismatch with references or severe frame-to-frame inconsistency.
\textbf{- 1:} Complete Failure. Generated content bears no resemblance to the reference images or is completely inconsistent. \\
\textbf{2. The second score: Prompt Following} \\
\textbf{Objective:} Assess how accurately the video follows the text instruction (scene composition, actions, camera movements, timing, etc.). \\
\textbf{- 5:} Perfect Alignment. Every detail of the instruction is perfectly rendered.
\textbf{- 4:} Good Alignment. The core semantics of the prompt are successfully captured.
\textbf{- 3:} Partial Alignment. The model captures the main idea but ignores specific nuances.
\textbf{- 2:} Weak Alignment. The video mostly ignores the prompt.
\textbf{- 1:} No Alignment. The video content has no relevance to the instruction. \\
\textbf{3. The third score: Visual Quality} \\
\textbf{Objective:} Evaluate the general aesthetic quality, temporal consistency, and motion smoothness. \\
\textbf{- 5:} Excellent. Exceptional visual aesthetics and perfect temporal consistency.
\textbf{- 4:} Good. High overall quality with only minor artifacts.
\textbf{- 3:} Fair. Acceptable but noticeably flawed.
\textbf{- 2:} Poor. Significantly degraded quality.
\textbf{- 1:} Unacceptable. Completely collapsed visual integrity. \\

\textbf{Example Response Format:}
You are required to return a dictionary structured as follows: \{"Subject Consistency": [A number from 1 to 5], "Prompt Following": [A number from 1 to 5], "Visual Quality": [A number from 1 to 5]\}. \\

The instruction is: \textless edit\_prompt\textgreater \\
Below are the reference images and generated video:

\end{AcademicBox}
\vspace{-1em}
\caption{Evaluation prompt template for multi-reference-image-to-video generation.\looseness=-1}
\label{fig:bench_multiref}
\end{figure}

\end{document}